\pdfoutput=1

\documentclass[11pt]{article}

\usepackage[preprint]{acl}

\usepackage{times}
\usepackage{latexsym}
\usepackage{multicol}
\usepackage{enumitem}

\usepackage{graphicx}
\usepackage{booktabs}
\usepackage{float}
\usepackage{stfloats}
\usepackage{amsmath}
\usepackage{multirow}
\usepackage{capt-of}
\usepackage{hyperref}
\usepackage{caption}
\usepackage{subcaption}

\usepackage{titletoc}

\usepackage[T1]{fontenc}

\usepackage[utf8]{inputenc}

\usepackage{microtype}

\usepackage{inconsolata}

\title{Instruction Fine-Tuning: Does Prompt Loss Matter?}

\author{Mathew Huerta-Enochian$^1$\thanks{Currently affiliated with the University of Oregon, research conducted while working at EQ4ALL.} \and Seung Yong Ko$^2$ \\
  EQ4ALL \\
  11 Nonhyeon-ro 76-gil, Gangnam-gu, Seoul \\
  $^1$\texttt{mhuerate@uoregon.edu} \\ $^2$\texttt{stephenko@eq4all.co.kr} \\
}

\begin{document}
\maketitle
\begin{abstract}
We present a novel study analyzing the effects of various prompt loss token weights (PLW) for supervised instruction fine-tuning (SIFT).
While prompt-masking (PLW $=0$) is common for SIFT, some fine-tuning APIs support fractional PLWs and suggest that using a small non-zero PLW can help stabilize learning when fine-tuning on short-completion data.
However, there has never been a study confirming this claim, and OpenAI, a major cloud-based SIFT provider, recently removed this parameter from their fine-tuning API.
We found that performance of models fine-tuned on short-completion data had a statistically-significant negative quadratic relationship with PLW.
Using small values ($0.01-0.5$) of PLW produced better results on multiple-choice and short-generation benchmarks (outperforming models fine-tuned on long-completion data) while large values ($\approx 1.0$) of PLW produced better results on long-generation benchmarks.
We explained this effect and verified its importance through additional experiments.
This research serves as a warning to API providers about the importance of providing a PLW parameter for SIFT.
\end{abstract}

\section{Introduction}\label{sec:intro}

Recent research in language modeling has made huge advances in training instruction-following agents.
Both supervised fine-tuning (SFT) and reinforcement learning (RL) have been employed to much success.
However, our understanding of optimal hyperparameters and standards of practice (SOPs) have been slow to catch up.
This research contributes to supervised instruction fine-tuning (SIFT) SOPs via an in-depth analysis of a single training hyperparameter: prompt loss weight (PLW).

While training, model parameters are updated by optimizing for next-token maximal likelihood classification.
Most open sourced solutions for SIFT either mask the prompt loss (for prefix language modeling) or use the entire sequence loss (for full language modeling) while some API providers support an explicit PLW parameter that allows users to apply fractional PLW during SIFT.
The commonly-held notion is that fractional PLW helps stabilize learning when fine-tuning on data with short outputs.
Recently, however, OpenAI quietly removed support for their \texttt{prompt\_loss\_weight} parameter.

The reason for the removal is unknown.
Furthermore, to our knowledge, there has never been a proper study on the effects of PLW.\footnote{Note that this research used a general implementation of PLW. OpenAI's former \texttt{prompt\_loss\_weight} implementation could not be tested directly.}

We make the following contributions:
\begin{itemize}
    \item We showed that PLW has a significant relationship with model performance when fine-tuning on \emph{short-completion} data.
    \item We showed that this relationship is due to a combination of regularizing effects and not the accepted explanation of increased training stability.
    \item We provided evidence that PLW cannot be replaced by other common regularizers alone and that PLW is important for strong performance on short-generation downstream tasks.
    \item We verified that PLW can be safely ignored when fine-tuning on \emph{long-completion} data.
\end{itemize}

We provide relevant background and hypotheses in section~\ref{sec:prelim} and \ref{sec:hyp}, respectively.
The main regression experiment is presented in sections \ref{sec:method} and \ref{sec:results}.
Supplemental experiments further validating our claims are presented in section~\ref{sec:supp}.
We present conclusions in section~\ref{sec:conclusion} followed by several appendices for additional analysis and discussion.

\section{Background}
\label{sec:prelim}

\subsection{Definitions}
\label{sec:def}

We define \emph{instruction data} as one or many instances of structured text data, each containing an instruction, an optional input, and a target output text.
We will use the term \emph{prompt} to refer to the concatenation of the instruction and input (if it exists) and the term \emph{completion} to refer to the target output.
The goal of SIFT is to fine-tune a model to generate an appropriate completion for a given prompt.

We define the \emph{generation ratio} $R_g$ as the ratio of completion length to prompt length (also referred to as the completion-prompt ratio).
We then divide instruction data into two broad categories.
Data with $R_g < 1$ are \emph{short-completion} data, and data with $R_g \ge 1$ are \emph{long-completion} data.
When applied to an entire dataset, we take $R_g$ to be the mean completion-prompt ratio.

\subsection{Relevant Research and Libraries}
\label{sec:literature}

HuggingFace's Transformers library \cite{Wolf_Transformers_State-of-the-Art_Natural_2020}, the de facto library for training LLMs, allows users to mask select tokens when calculating token classification loss.
In Transformers, weights for next-token prediction loss is therefore binary\textemdash either token loss is masked (PLW $=0$) or it is unmasked (PLW $=1$).

As mentioned in section~\ref{sec:intro}, OpenAI officially removed support for a \texttt{prompt\_loss\_weight} parameter in their fine-tuning API as part of the v1 fine\_tune API deprecation in early January, 2024.
This \texttt{prompt\_loss\_weight} parameter used a default value of $0.01$ with the following parameter explanation: \emph{
``This controls how much the model tries to learn to generate the prompt (as compared to the completion which always has a weight of 1.0), and can add a stabilizing effect to training when completions are short.
If prompts are extremely long (relative to completions), it may make sense to reduce this weight so as to avoid over-prioritizing learning the prompt.''
}

Though we could not find a study validating OpenAI's claim or any literature that presents an analysis of PLW, we found several studies that reported using this parameter.
Though they do not provide their reasoning, \citet{kozachek2023investigating} reported that they fine-tuned GPT-3 with a \texttt{prompt\_loss\_weight} of $0.1$.
\citet{dodgson2023establishing} reported using the default value of $0.01$ when fine-tuning GPT models.
\citet{wang-2023-self-instruct} reported that a PLW of $0$ performed best for them when working on the Self-Instruct framework.
Interestingly, \citet{wutschitz2023rethinking} reported hyperparameter search results for next-sentence-prediction on Elsevier data using PLWs of $0.1$ and $0.5$ and found $0.5$ to give the best results.
Similar to OpenAI's deprecated API, BLoomAI's API supports a \texttt{prompt\_loss\_weight} parameter with a default value of 0.01.

\begin{table*}[tbh]
    \centering\small
    \begin{tabular}{lccccc}
        \toprule
        &\multicolumn{3}{c}{Mean (Std) Tokens}& & \\
        \cmidrule(lr){2-4}
        \multicolumn{1}{c}{Dataset} & Instruction & Input & Completion & Total Tokens & $R_g$ \\
        \midrule
        AlpacaData & 13.40 (4.97) & 6.25 (14.38) & 64.51 (64.85) & 4,376,482 & 3.27 \\
        AlpacaDataCleaned & 14.21 (9.95) & 6.42 (17.65) & 162.74 (150.89) & 9,490,837 & 7.83 \\
        AlpacaDataShort & 16.93 (13.10) & 162.34 (151.69) & 14.62 (10.99) & 10,035,667 & 0.08 \\  
        UltraFeedback* & 184.47 (268.04) & 0 (-) & 327.37 (291.48) & 31,075,393 & 1.77 \\
        DatabricksDolly* & 18.65 (74.55) & 91.60 (250.66) & 91.14 (149.15) & 3,023,113 & 0.83 \\
        UltraFeedbackShort* & 94.09 (141.94) & 206.18 (211.03) & 55.03 (61.00) & 16,351,333 & 0.18 \\
        DatabricksDollyShort* & 18.83 (75.29) & 160.37 (270.17) & 28.79 (47.49) & 3,122,209 & 0.16 \\
        \bottomrule
    \end{tabular}
    \caption{
    Dataset statistics: 
    mean tokenized instruction, input, and completion sequence lengths (standard deviations in parentheses), 
    total token counts for each dataset, and the generation ratio $R_g$. \\
    * Used for supplemental experiments in section~\ref{sec:supp}.
    }
    \label{tab:data}
\end{table*}

\section{Hypotheses}
\label{sec:hyp}

Based on OpenAI's explanation for \texttt{prompt\_loss\_weight}, we expected that for SIFT with short-completion data and small values of PLW, there would be a positive relationship between PLW and downstream performance.
However, training a model to maximize next-token-prediction on prompt tokens should be most useful for generating instruction data, and over-prioritizing prompt token loss should have a negative influence on downstream performance.

Based on these assumptions, we would expect the two competing factors to result in a downward curved relationship between PLW and downstream performance.
Limiting PLW to the range of [0, 1], we postulate that there is a critical value $\lambda$ for PLW with $0 <= \lambda <= 1$.
For PLW less than $\lambda$, the positive effect dominates the negative effect and for values greater than $\lambda$, the negative effect dominates the positive effect.
If $\lambda= 0$, then PLW's contribution to model performance is strictly negative, and if $\lambda= 1$, then PLW contributes strictly positively to model performance.
Note that $\lambda$ would not be an intrinsic characteristic of the dataset, model architecture, or training algorithm.
Rather, it would depend on numerous factors and change for each task.

We then made two hypotheses to test the above relationship and performed regression analysis on three fine-tuning datasets spanning a range of $R_g$ values: 0.08, 3.27, and 7.83.

\noindent
\textbf{Null Hypothesis} ($\mathbf{H_0}$)
\emph{Prompt loss weight has no relationship with model performance.}

\noindent
\textbf{Alternative} ($\mathbf{H_1}$)
\emph{Prompt loss weight has a quadratic relationship with model performance.}

We used the standard $\alpha=0.05$ significance level, and we expected to reject $\mathbf{H_0}$ only for models trained on short-completion data.

\section{Methodology}\label{sec:method}

To evaluate the effect of PLW on downstream performance, we used a factorial design methodology and repeated the Alpaca experiment \cite{alpaca} with three experimental variables.
We tested ten discrete levels of PLW, two pre-trained language models (PTLMs), and three instruction fine-tuning datasets for a total of sixty experimental training runs and evaluated each run on thirteen benchmarks.

We used the original Alpaca code and Transformers library, only modified to add PLW.
Training was performed exactly as per the original Alpaca experiment, and we used the hyperparameters suggested by the authors, modifying only the three experimental parameters (PLW, PTLM, dataset) with each run.

We provide additional details for reproducibility in appendix~\ref{sec:app-reproduce} and will release our trained models on HuggingFace's Hub. 

\subsection{Prompt Loss Weight}

We limited our evaluation of PLW to factors in the range [0, 1], focusing on values close to zero: \\
\begin{align*}
    \text{PLW} \in & \{0.0, 5{\times}10^{-4}, 2.236{\times}10^{-3}, \\
    & 1{\times}10^{-2}, 2.463{\times}10^{-2}, 5{\times}10^{-2}, \\
    & 1{\times}10^{-1}, 2.463{\times}10^{-1}, 5{\times}10^{-1}, 1.0\}
\end{align*}
Note that PLW $=0.0$ is identical to the masking used in the original Alpaca project, and PLW $=1.0$ is equivalent to unmasked training.

For all analysis, we transformed our PLW values to be closer to uniform on the interval [0, 1] using a power function
\[
  f\colon \biggl\{\begin{array}{@{}r@{\;}l@{}}
    [0, 1] &\to [0, 1],\\
    v &\mapsto v^{p}
  \end{array}
\]
where the power $p = 0.30103$ was chosen semi-arbitrarily such that $f(0.1) = 0.5$.
We denote the transformed PLW values as \texttt{w\textsubscript{p}}

\subsection{Pre-Trained Language Model}

We fine-tune both LLaMA 1 7B \cite{touvron2023llama} to recreate the original Alpaca experiment and LLaMA 2 7B \cite{touvron2023llama2} to provide more relevant results.

\subsection{Fine-Tuning Dataset}

We ran all experiments with three datasets: AlpacaData (the instruction dataset from the original Alpaca experiment), AlpacaDataCleaned \cite{alpacadatacleaned}, and AlpacaDataShort.

AlpacaDataCleaned is a cleaned and curated version of AlpacaData that has recently been combined with data from the GPT4 LLM dataset \cite{peng2023instruction}.
Cleaning is noted as ongoing and includes fixes for the following issues in AlpacaData: hallucinations, merged instructions, empty outputs, empty code examples, instructions to generate images, N/A outputs, inconsistent input fields, wrong answers, nonsensical instructions, and extraneous control characters.

We generated AlpacaDataShort from AlpacaDataCleaned by rephrasing long-completion instances as prompt-prediction task, a process we denote as \emph{prompt inversion}.
See Appendix~\ref{sec:app-short} for more on prompt inversion.

Descriptive statistics for these each dataset are presented in table~\ref{tab:data}.
Note that AlpacaDataCleaned is strongly long-completion with an $R_g$ of $7.83$ while AlpacaDataShort is short-completion with an $R_g$ of $0.082$.

\subsection{Performance Evaluation}\label{sec:eval}

\begin{table}[t]
  \centering\small
  \begin{tabular}{lcccc}
    \toprule
    \multicolumn{1}{c}{Task} & V. & Shots & Split & Type \\
    \midrule
    \rule{0pt}{3ex}
    ARC Challenge* & 0 & 25 & Test & MC\\
    \rule{0pt}{2ex}
    PIQA & 0 & 0 & Val & MC\\
    \rule{0pt}{2ex}
    TruthfulQA-MC2* & 1 & 6\dag & Val & MC \\
    \rule{0pt}{2ex}
    WinoGrande* & 0 & 5 & Val & MC \\
    \rule{0pt}{2ex}
    TruthfulQA-Gen & 1 & 6\dag & Val & G\textsubscript{S} \\
    \rule{0pt}{2ex}
    WMT14 En$\rightarrow$Fr & 1 & 0 & Val+Test & G\textsubscript{S} \\
    \rule{0pt}{2ex}
    WMT14 Fr$\rightarrow$En & 1 & 0 & Val+Test & G\textsubscript{S} \\
    \rule{0pt}{2ex}
    WMT16 En$\rightarrow$De & 1 & 0 & Val+Test & G\textsubscript{S} \\
    \rule{0pt}{2ex}
    WMT16 De$\rightarrow$En & 1 & 0 & Val+Test & G\textsubscript{S} \\
    \rule{0pt}{2ex}
    WMT16 En$\rightarrow$Ro & 1 & 0 & Val+Test & G\textsubscript{S} \\
    \rule{0pt}{2ex}
    WMT16 Ro$\rightarrow$En & 1 & 0 & Val+Test & G\textsubscript{S} \\
    \rule{0pt}{2ex}
    AlpacaEval (Mixtral) & 1 & 1 & Test & G\textsubscript{L} \\
    \rule{0pt}{2ex}
    PandaLM & 1 & 0 & Test & G\textsubscript{L} \\
    \bottomrule
  \end{tabular}
  \caption{
  Evaluation benchmarks.
  Validation splits were used when test splits were unavailable, and validation and test splits were combined for noisy benchmarks.
  Benchmark completion type is noted here as ``MC'' for multiple choice, ``G\textsubscript{S}'' for short generation, and ``G\textsubscript{L}'' for long-generation. \\
  *Task used in HuggingFace's Open LLM Leaderboard. \\
  \dag This benchmark was calculated with \texttt{num\_fewshot} $=0$ but uses a built-in minimum of 6 shots.
  }
  \label{tab:bench}
\end{table}

We evaluated each model on thirteen instruction benchmarks covering multiple choice and text generation tasks.
We selected benchmarks that were relatively cheap to compute and covered a range of tasks.
We used three evaluation frameworks: EleutherAI's Language Model Evaluation Harness (EEH) \cite{eval-harness}, AlpacaEval 1 \cite{alpaca_eval}, and PandaLM \cite{pandalm, wang2024pandalm}.
See table~\ref{tab:bench} for details on benchmark tasks.

Eleven benchmarks were run using EEH.
Four of these were multiple choice tasks:
ARC Challenge \cite{Clark2018ThinkYh},
PIQA \cite{Bisk2020},
TruthfulQA-MC2 \cite{lin-etal-2022-truthfulqa}, and
WinoGrande \cite{sakaguchi2019winogrande}.
Seven were short generation tasks:
TruthfulQA-Gen \cite{lin-etal-2022-truthfulqa}
and six WMT14 and WMT16 translation benchmarks \cite{bojar-etal-2014-findings, bojar-etal-2016-findings}, limited to four languages the PTLMs saw during pretraining (English, French, German, and Romanian).
For WMT benchmarks, we used the zero-shot instruction ``\texttt{Translate the following from <src\_lang> to <tgt\_lang>}'' and evaluated over both the validation and test sets to reduce variance.

Long-generation performance was evaluated using AlpacaEval 1 and PandaLM, which are both LLM-as-a-judge frameworks.
The default auto-evaluator for AlpacaEval 1 is GPT4, but using paid APIs would beyond the scope of this research, so we used Mixtral 8X7B \cite{jiang2024mixtral} as an auto-evaluator.
Mixtral performed the best of all open-source LLMs that we tested on AlpacaEval's evaluator test dataset \cite{dubois2023alpacafarm}, with 64.9\% agreement with human evaluators.
For reference, Claude \cite{claudeai} has 65.3\% and GPT4 has 70.99\% human agreement.

\section{Results and Discussion}
\label{sec:results}

\begin{figure*}[th!]
    \centering
    \begin{subfigure}[b]{.32\linewidth}
        \includegraphics[width=\linewidth]{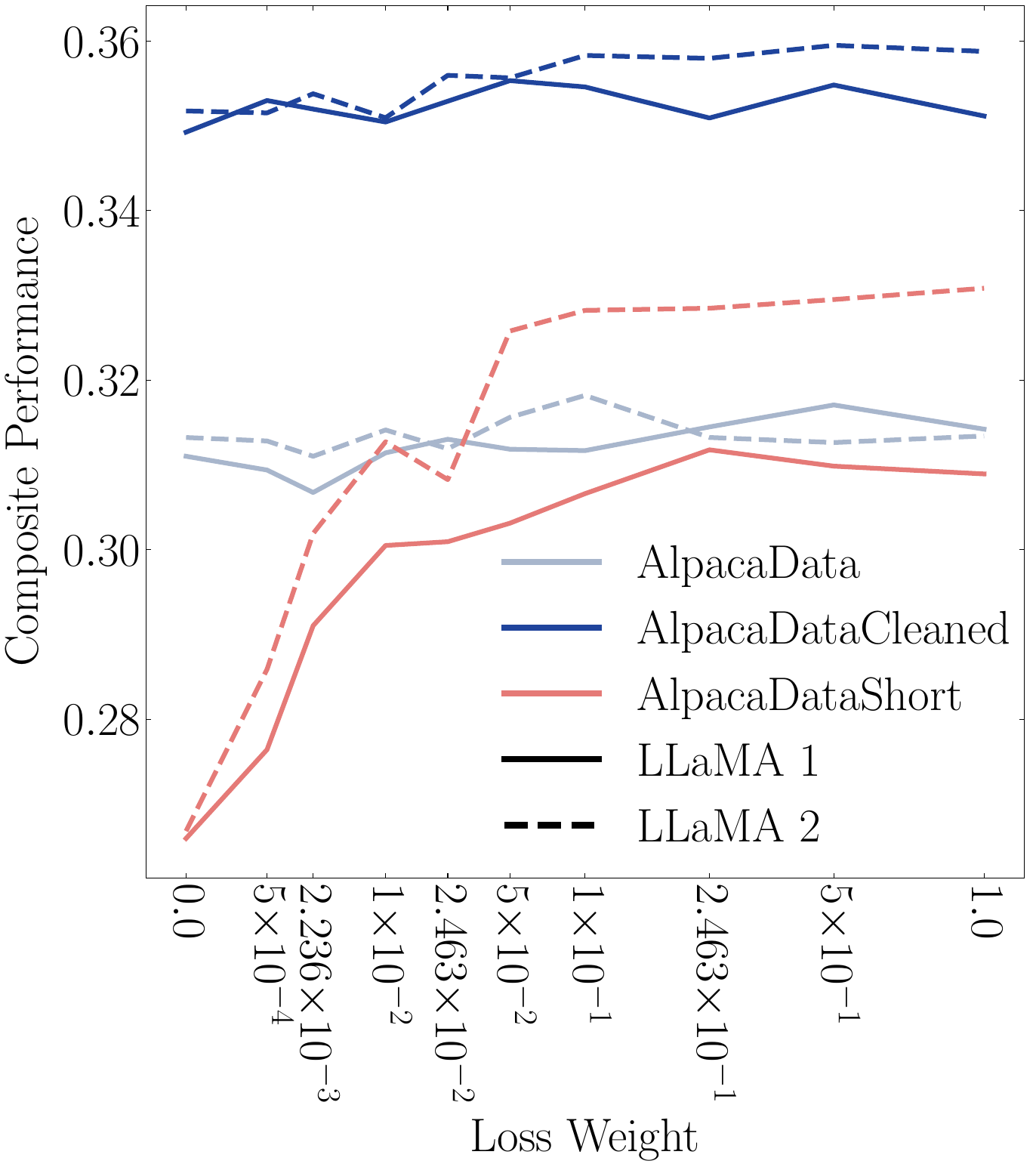}
        \vspace{-0.6cm}
        \caption{Simple Aggregate}
    \end{subfigure}
    \hspace{2cm}
    \begin{subfigure}[b]{.32\linewidth}
        \includegraphics[width=\linewidth]{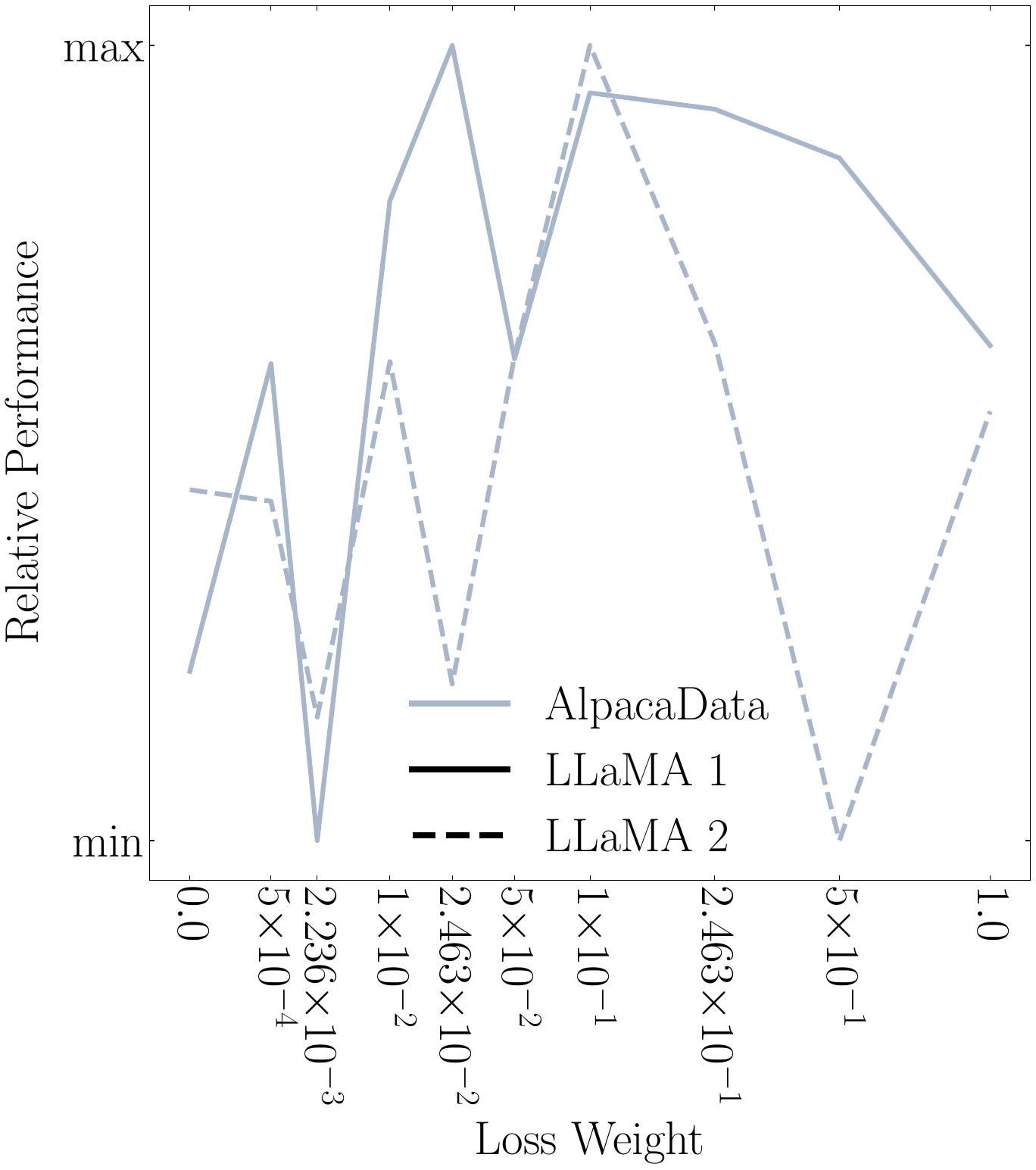}
        \vspace{-0.6cm}
        \caption{AlpacaData Relative Aggregate}
    \end{subfigure} \\
    \vspace{0.5cm}
    \begin{subfigure}[b]{.32\linewidth}
        \includegraphics[width=\linewidth]{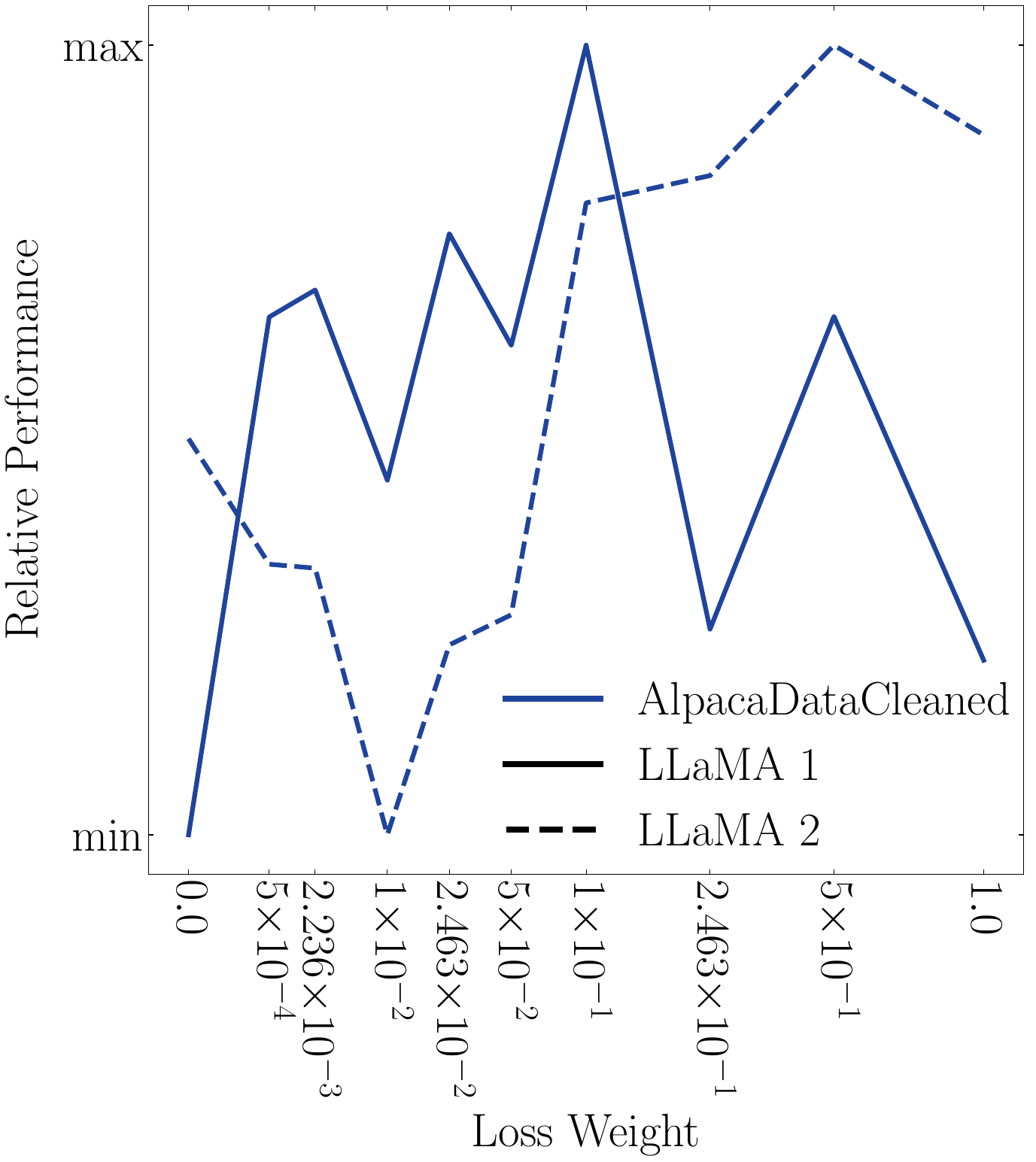}
        \vspace{-0.6cm}
        \caption{AlpacaDataClean Relative Aggregate}
    \end{subfigure}
    \hspace{2cm}
    \begin{subfigure}[b]{.32\linewidth}
        \includegraphics[width=\linewidth]{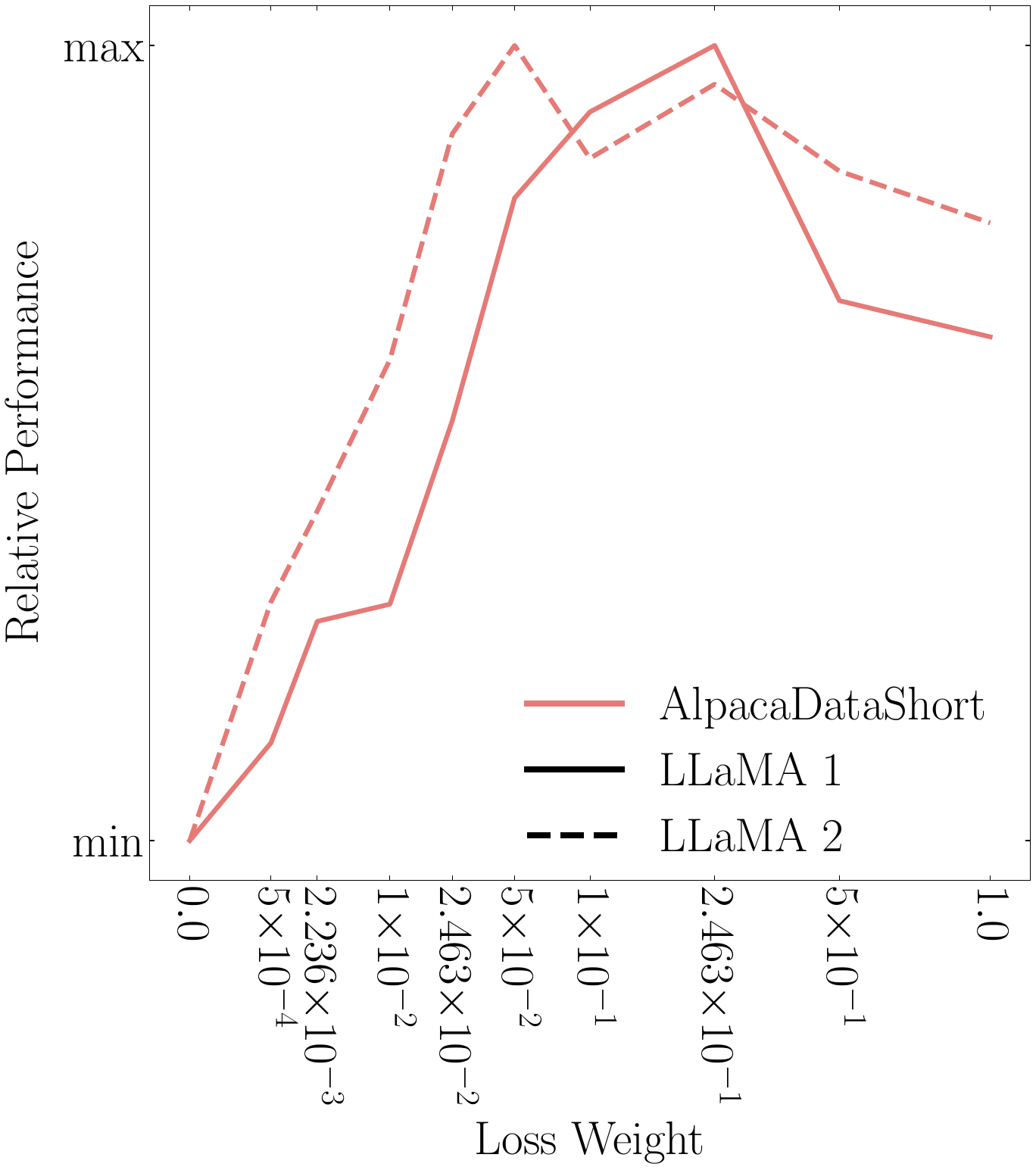}
        \vspace{-0.6cm}
        \caption{AlpacaDataShort Relative Aggregate}
        \label{fig:short-rel-agg}
    \end{subfigure} \\
    \caption{
    Performance by transformed PLW.
    \textbf{(a)} A simple performance aggregate score (the unweighted mean of benchmark scores).
    \textbf{(b)}, \textbf{(c)}, \textbf{(d)} Relative aggregate performance scores where scores per task for each task and group are min-max scaled to show common trends, regardless of scale.
    Note that aggregate scores for only the AlpacaDataShort models show a relationship with transformed PLW.
    Best viewed in color.
    }
    \label{fig:taskperformance}
    \label{fig:composite}
\end{figure*}

A visualization of the simple and min-max scaled relative performance aggregates by \texttt{w\textsubscript{p}} and dataset is presented in figure~\ref{fig:composite}.
Note that all analysis was performed using the relative aggregate.
The simple aggregate is dominated by large performance changes on a few benchmarks and is included only for completeness.

\subsection{Performance Trends}

There are several qualitative performance trends that are of interest.
For more thorough discussion and task-specific benchmark plots, see Appendix~\ref{sec:app-taskperf}.

A group of four benchmarks (Arc Challenge, PIQA, TruthfulQA-Gen, and WinoGrande) clearly show the expected negative quadratic relationship between \texttt{w\textsubscript{p}} and performance of AlpacaDataShort models, with optimal PLW somewhere in $0 < \texttt{w\textsubscript{p}} < 1$.
Notably, AlpacaDataShort models outperform AlpacaData and AlpacaDataCleaned on this group given optimal PLW tuning.
On the long-generation benchmarks, however, AlpacaDataShort models show a steadily-increasing trend, with optimal PLW near $\texttt{w\textsubscript{p}} = 1$.
For these seven benchmarks, AlpacaDataShort models fine-tuned with prompt loss masking (\texttt{w\textsubscript{p}}$=0$) almost always produced the worst scores.

Maximal \texttt{w\textsubscript{p}}-based performance increase was around twenty percentage points for both long-generation benchmarks and less than two percentage points for short-generation and multiple choice benchmarks.
This difference in scale and the above difference in optimal PLW shows that the relationship between PLW and model performance is strongly dependent on the benchmark task.

Clear qualitative performance trends for the AlpacaData- and AlpacaDataCleaned-trained models and for any model evaluated on the six translation benchmarks could not be identified.

\subsection{Regression}
\label{sec:regression}

For each data group, we fit a generalized linear mixed model (GLMM) with the relative aggregate benchmark scores as the response variable.

We expected a quadratic relationship between the score and \texttt{w\textsubscript{p}}, so we included a second order polynomial of \texttt{w\textsubscript{p}} as a fixed effect.
Furthermore, we knew that the PLW-performance relationship varies by benchmark and since scores were min-max normalized over each benchmark, we used a random slope (and no intercept) with respect to benchmark.
Since we did not min-max normalize over PTLM groups and since we saw consistent improvement when using LLaMA 2, we modeled a random intercept for PTLM. This resulted in the following equation that we fit with the \textbf{\textsf{R}} library \texttt{glmmTMB}: \\
$\texttt{score} \sim \texttt{pol(w\textsubscript{p},2)} + (0{+}\texttt{pol(w\textsubscript{p},2)}|\texttt{b}) + (1|\texttt{m})$
where \texttt{score} is the min-max transformed scores, \texttt{b} is the benchmark task factor, and \texttt{m} is the PTLM factor.
Since \texttt{score} is bounded and thus introduced heteroskedasticity, we used a beta distribution as the conditional distribution of the response variable.
Model fit was evaluated with the \texttt{DHARMa} library and \texttt{glmmTMB}'s \texttt{Anova} method.

\begin{table}[t!]
    \centering\small
    \begin{tabular}{c@{\hspace{5pt}}cc@{\hspace{5pt}}cc}
        \toprule
        & & \multicolumn{2}{c}{Coeff} & \\
        \cmidrule(lr){3-4}
        & P-Value & \texttt{\normalsize w\textsubscript{p}} & \texttt{\normalsize w\textsubscript{p}}\textsuperscript{2} & (Int) \\
        \midrule
        AlpacaData & 0.237 & 1.185 & -0.917 & (-0.131) \\
        AlpacaDataCleaned & 0.0861 & 1.238 & -0.812 & (-0.231) \\
        \textbf{AlpacaDataShort} & \textbf{<0.001} & \textbf{5.590} & \textbf{-4.284} & (-1.043) \\
        \bottomrule
    \end{tabular}
    \caption{
    \texttt{w\textsubscript{p}} p-values and coefficients by training dataset.
    Statistically significant results are in \textbf{bold}.
    Note that though convergence warnings were raised for regression on both AlpacaData and AlpacaDataCleaned, coefficient and p-value scores are reported for completeness.
    }
    \label{tab:regression}
\end{table}

\begin{figure*}[th!]
    \centering

    \begin{subfigure}[b]{.9\linewidth}
        \includegraphics[width=\linewidth]{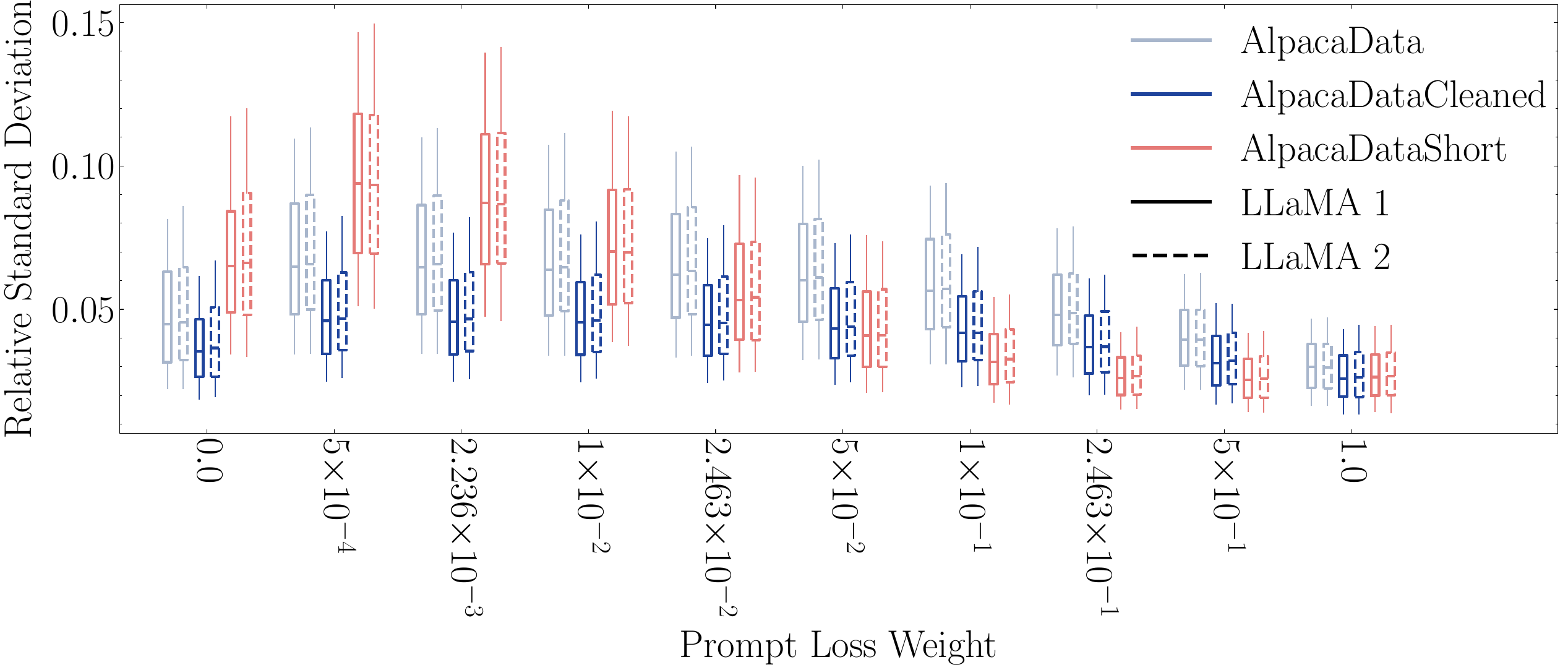}
        \vspace{-0.6cm}
        \caption{Training Loss Stability}
        \label{fig:rsd}
    \end{subfigure} \\
    \vspace{3mm}
    \begin{subfigure}[b]{.3\linewidth}
        \includegraphics[width=\linewidth]{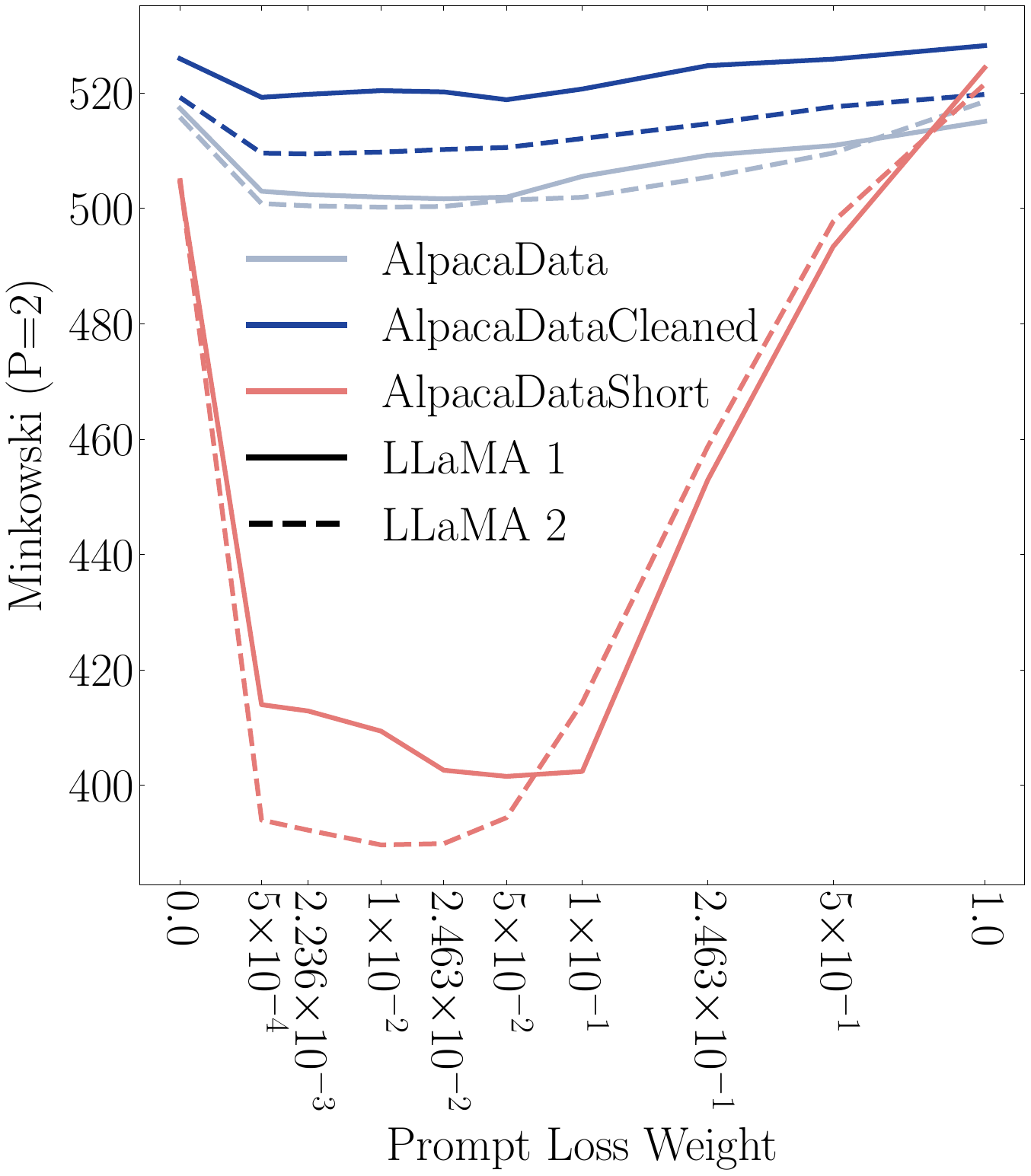}
        \vspace{-0.6cm}
        \caption{Weight Distance}
        \label{fig:weightdist}
    \end{subfigure}
    \hfill
    \begin{subfigure}[b]{.292\linewidth}
        \includegraphics[width=\linewidth]{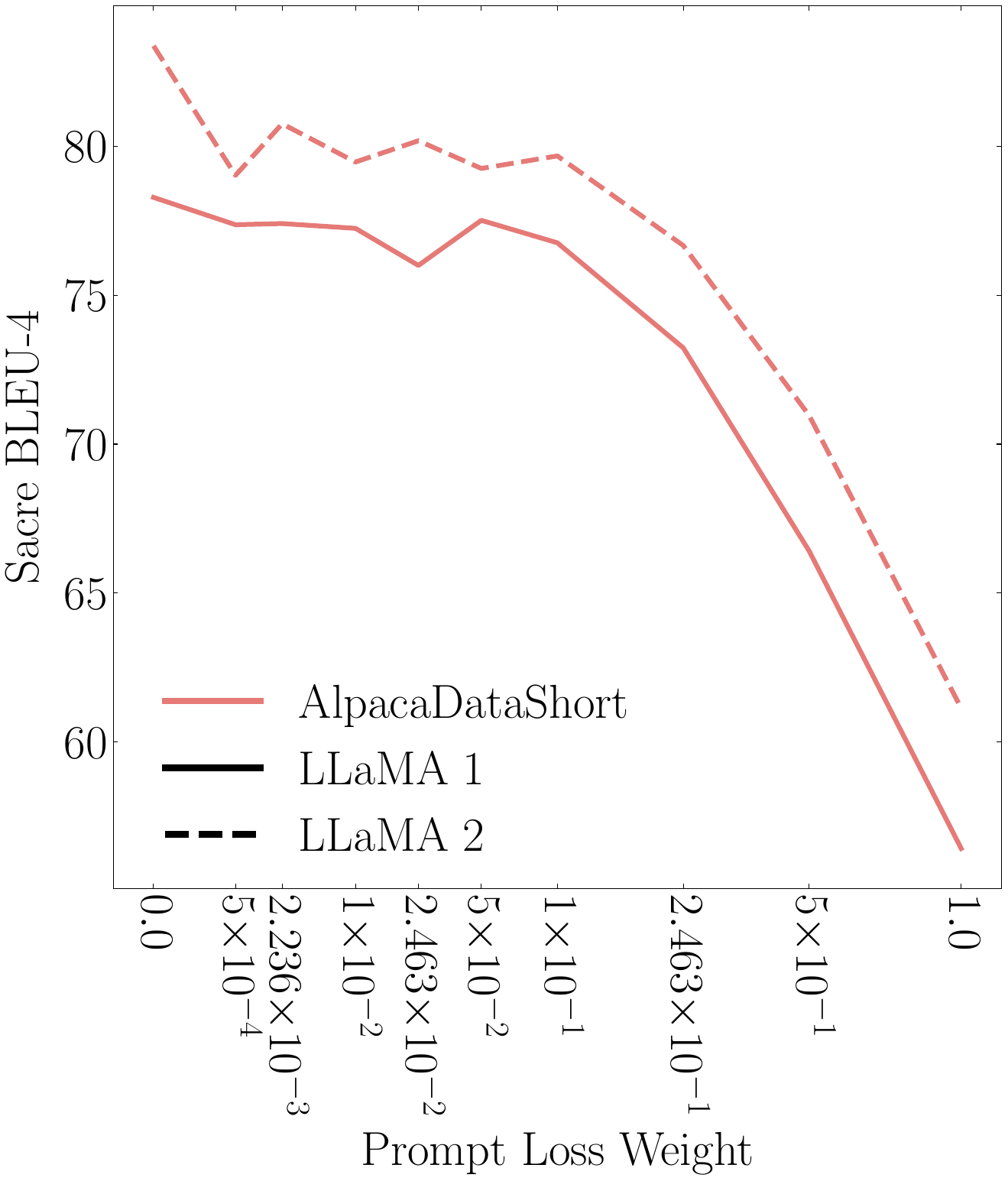}
        \vspace{-0.6cm}
        \caption{Train Data Memorization}
    \end{subfigure}
    \hfill
    \begin{subfigure}[b]{.33\linewidth}
        \includegraphics[width=\linewidth]{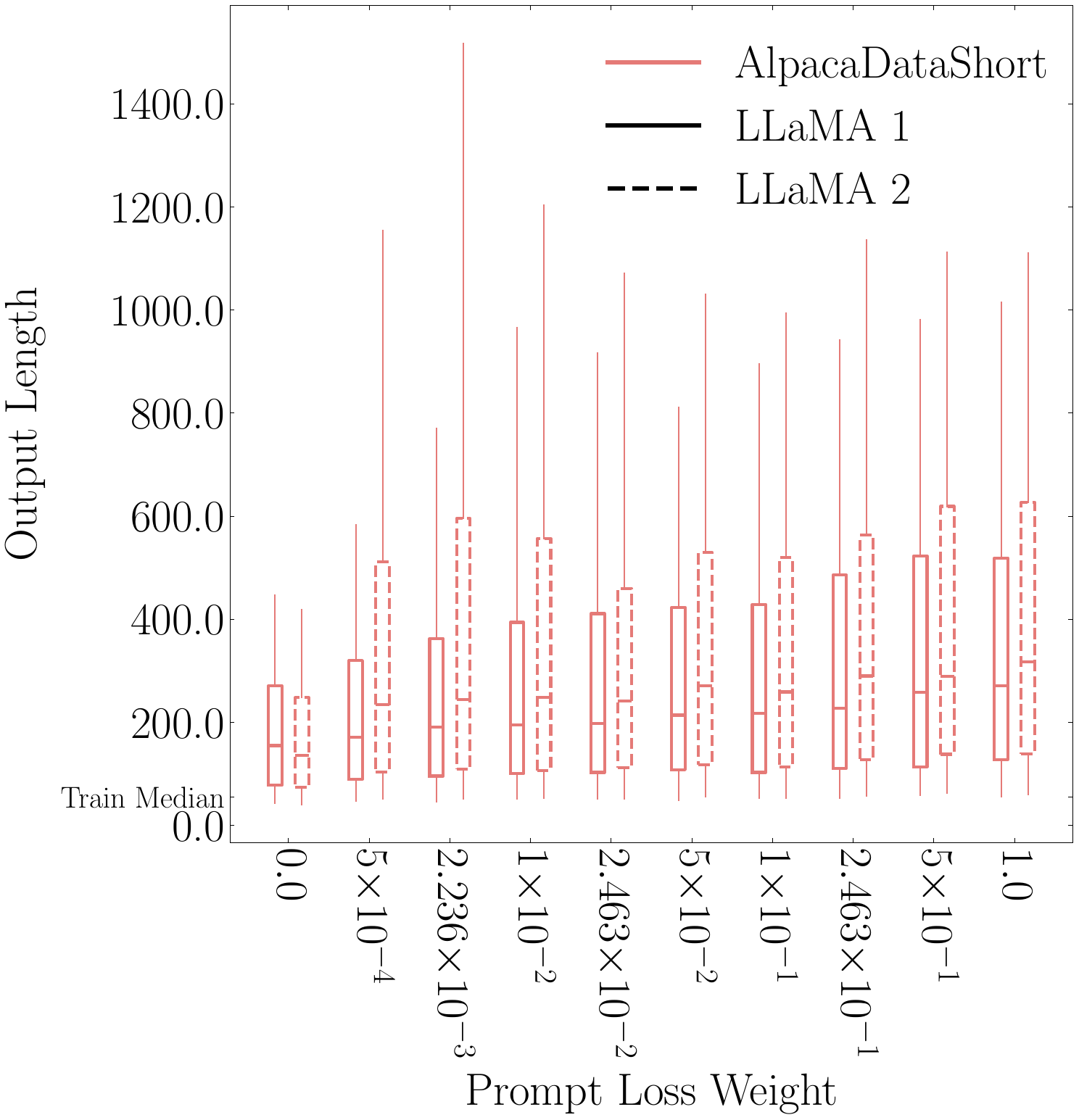}
        \vspace{-0.6cm}
        \caption{AE Generation Length}
        \label{fig:aelength}
    \end{subfigure}\\

    \caption{
    Analysis of causal mechanism.
    Boxplots use the 0.25, 0.5, and 0.75 quantiles with whiskers at 0.09 and 0.91 quantiles.
    Best viewed in color.
    \textbf{(a) Training Loss Stability}: Relative Standard Deviation (RSD) of five-step training loss windows show increase instability for small (non-zero) PLWs. 
    \textbf{(b) Weight Distance}: Distance between learned weights and PTLM weights is smaller for small (non-zero) PLWs.
    \textbf{(c) Train Data Memorization}: Completion Sacre BLEU scores on training data prompts as an indicator for overfitting.
    \textbf{(d) AE Generation Length}: Generation lengths on the Alpaca Eval test set for varying PLW values.
    }
    \label{fig:explain}
\end{figure*}

P-values and coefficients are presented in table~\ref{tab:regression}.
Regression on both AlpacaData and AlpacaDataCleaned produced convergence warnings and appropriate models could not be adequately fit.
We tried reducing the complexity of the model, but no significant relationship with \texttt{w\textsubscript{p}} could be found.
However, the model fit on AlpacaDataShort converged and passed residual normallity, homoscedasticity, and other checks for soundness.
For the AlpacaDataShort case, min-max transformed performance showed a statistically significant negative quadratic relationship with \texttt{w\textsubscript{p}} at our target $\alpha=0.05$ significance level.

This means that while we could not reject the null hypothesis for the AlpacaData and AlpacaDataCleaned scenarios, for the AlpacaDataShort scenario, there was sufficient evidence to reject the null hypothesis in favor of the alternative hypothesis \emph{H\textsubscript{1}}.

Using the fixed effect coefficients, we can predict the critical PLW value $\lambda$ for AlpacaDataShort fine-tuning that maximizes the min-max transformed benchmark scores.
The coefficients for \texttt{w\textsubscript{p}}\textsuperscript{2}, \texttt{w\textsubscript{p}}, and the intercept were -4.28, 5.590, -1.043, respectively.
We can rewrite this relationship as: \\
\centerline{$\texttt{score}=-4.284(\text{\texttt{w\textsubscript{p}}} - 0.652)^2 + 0.781$,} \\
which has a global maximum at $\texttt{w\textsubscript{p}}=0.652$.
Reversing the power transformation yields a critical value for PLW at $\lambda = 0.242$.
We verified that this predicted $\lambda$ overlaps with the visualized maximum value range for the relative aggregate in figure~\ref{fig:short-rel-agg}.

\subsection{Causal Mechanism \& Interpretation}

\subsubsection{Training Stability}

To identify possible causal mechanisms, we first investigated the effects of PLW on training stability by analyzing training loss relative standard deviation (RSD) over five-step windows.
See figure~\ref{fig:rsd} for a boxplot of mean RSD for each model.
For all dataset and PTLM factors, increasing PLW from zero led to a sharp increase in mean RSD and then a slow decrease to a minimum mean RSD at PLW $=1$.
There is no obvious explanation for why training loss RSD would increase for small PLW before decreasing for large PLW.

If training loss stability was the primary factor in improved performance, we would expect RSD to be lowest for PLW between $0.01$ and $0.5$ (or even between $0.01$ and $0.1$ based on short-generation benchmarks) and for performance at PLW $=0$ to be similar with performance at PLW $=0.01$ since mean RSD at these values are similar.
However, mean RSD drops by a factor of two across the PLW $\in [0.01, 0.1]$ range, and performance at PLW $=0.01$ is significantly higher than the masked prompt loss scenario.
Training loss mean RSD is lowest at PLW $=1$, but performance on ARC Challenge, PIQA, WinoGrande, and TruthfulQA-Gen show clear decreasing trends at this value.
Furthermore, the three tasks showing positive trends at PLW $=1$ cannot be adequately explained by this factor since performance increases regardless of loss stability.

There is likely either a tradeoff between training loss stability and some other factors that affects model performance or model loss stability is not an important factor.
Considering that both AlpacaData and AlpacaDataCleaned models also showed a negative quadratic trend for mean loss RSD, we tentatively concluded that loss stability is not the driving factor for the modeled relationship.

\subsubsection{Weight Regularization}

We then checked if PLW was providing regularization to the weight update step, possibly improving performance by keeping weights close to the PTLM.
See figure~\ref{fig:weightdist} for a visualization of weight distance from PTLM.
Interestingly, for AlpacaDataShort, fine-tuned weights were closer to those of the PTLM for small values of PLW but were much farther for PLW $< 0.0005$ and PLW $> 0.1$.
We would expect weights to change more when loss is erratic, but the range of PLW values that better preserved PTLM weights was similar to the range of increased training loss RSD.
This is an interesting result, and we conclude that PTLM model weights were better preserved for small non-zero PLW \emph{despite} high loss instability.

\subsubsection{Data Memorization}

We next explored how PLW affected training data memorization.
We sampled 10,000 unique prompts from the AlpacaDataShort training set, generated completions from each prompt, and calculated corpus BLEU-4 scores.
We found that for PLW from 0.0 to around 0.1, models memorized most of the training data, consistently scoring near 80 corpus BLEU.
Corpus BLEU then decreased as PLW increased from 0.1.
We also analyzed generation length on the AlpacaEval 1 test set, which showed a generally increasing trend with PLW.\footnote{
Note that recent work \cite{alpaca_eval, dubois2024length} has shown that AlpacaEval 1 has a preference for long generations, and we argue that improved AlpacaEval scores are not simply due to longer generations.
First, while AlpacaEval performance showed a nearly strictly-increasing relationship with PLW, generation length did not.
Second, we used Mixtral as the auto-evaluator which showed a much lower length preference than the default evaluator (0.63 and 0.75, respectively).
}
Since AlpacaDataShort is dominated by short-completion instances, we concluded that non-zero PLW decreases overfitting by allowing the model to learn generation patterns from the prompt without negatively impacting instruction-completion alignment.

\subsubsection{Interpretation}

Based on the above analysis, we suggest that the causal mechanisms between PLW and downstream performance of models fine-tuned on short-completion data are
\begin{enumerate}
    \item preservation of PTLM weights for small PLW and
    \item reduced overfitting (and increased generation length) for large PLW.
\end{enumerate}
A tradeoff between these two mechanisms would explain the positive trend seen in the AlpacaEval and PandaLM benchmarks and the negative quadratic relationship in several of the other benchmarks.

\section{Supplemental Experiments}\label{sec:supp}
In this section, we present supplemental experiments that suggest that PLW cannot be replaced by alternative regularization techniques for SIFT and that the effects of PLW extend to other short-completion datasets.
Since the translation benchmarks showed high levels of noise and unclear correlation with PLW in the main experiment, they were not included in supplemental experiments.

\subsection{Alternative Regularizers}\label{sec:supp-reg}

\begin{table}[!tb]
    \centering\small
    \begin{tabular}{clccc}
        \toprule
        \multirow{2}{*}{LLaMA} & \multirow{2}{*}{Regularization} & \multicolumn{2}{c}{Aggregate} \\
        && Simple & Relative \\
        \midrule
        \multirow{5}{*}{1} & Weight Decay & \textbf{0.507} & 0.795 \\
        & Minkowski Metric & 0.505 & 0.812 \\
        & Dropout & 0.500 & 0.783 \\
        & Label Smoothing & 0.500 & 0.785 \\
        & PLW (Ours) & 0.506 & \textbf{0.855} \\
        \midrule
        \multirow{5}{*}{2} & Weight Decay & 0.537 & 0.773 \\
        & Minkowski Metric & 0.538 & 0.772 \\
        & Dropout & 0.541 & 0.805 \\
        & Label Smoothing & \textbf{0.538} & 0.837 \\
        & PLW (Ours) & 0.537 & \textbf{0.894} \\
        \bottomrule
    \end{tabular}
    \caption{
    Supplemental comparison of AlpacaDataShort models fine-tuned with various regularizers.
    High scores are in bold.
    As explained in section~\ref{sec:results}, the relative aggregate should be used for analysis, and the simple aggregate is provided for reference only.
    }
    \label{tab:short-reg}
\end{table}

To investigate if the effects of PLW on short-completion SIFT can be emulated with other common regularization techniques, we repeated the AlpacaDataShort training runs for PLW $=0$ and PLW $=1$ while applying various regularizers.
We tested weight decay, minimizing the Minkowski distance between PTLM weights and learned weights, dropout, and label smoothing.
See Appendix~\ref{sec:app-reg} for visualizations. 

Note that we do not compare PLW with KL-divergence.
While using KL-divergence as a regularizing loss is common for SFT in general and is used in RL-based LLM alignment \citep{stiennon2020learning,korbak2022rl,gao2023scaling}, collecting embeddings for any non-trivial dataset to be used for LLM SIFT presents a huge computation and memory overhead.

As can be seen in the results in table~\ref{tab:short-reg}, relative aggregate scores were higher for models fine-tuned with fractional PLWs.
This suggests that PLW provides a unique regularizing effect that cannot be easily replaced with other regularizers.

\subsection{Alternative Datasets}\label{sec:supp-data}

\begin{figure*}[tb!]
    \begin{subfigure}[b]{.32\linewidth}
        \label{fig:short-aug-norm}
        \vspace{-0.6cm}
        \includegraphics[width=\linewidth]{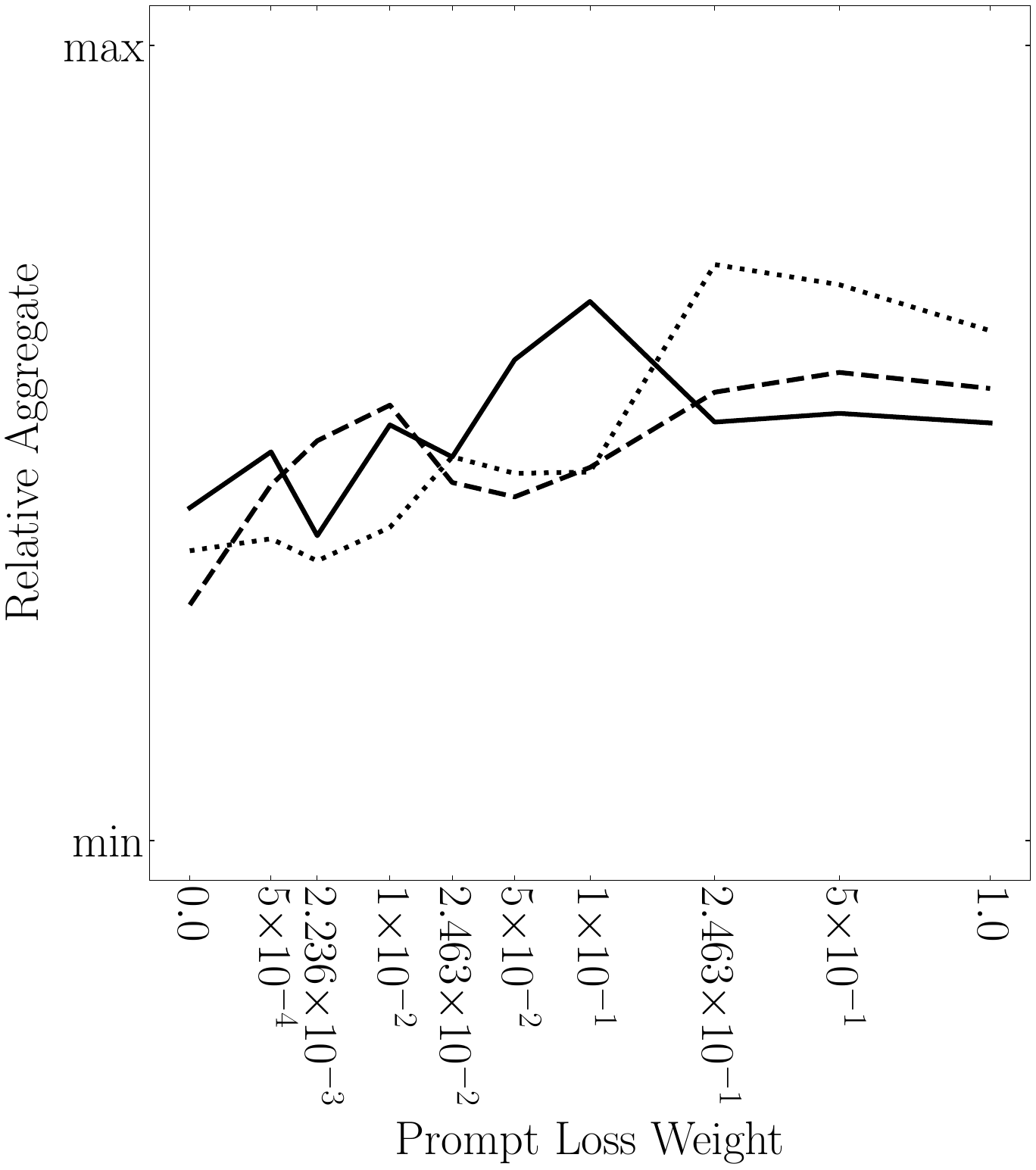}
        \caption{Original Datasets}
    \end{subfigure}
    \hspace{1cm}
    \begin{subfigure}[b]{.328\linewidth}
        \includegraphics[width=\linewidth]{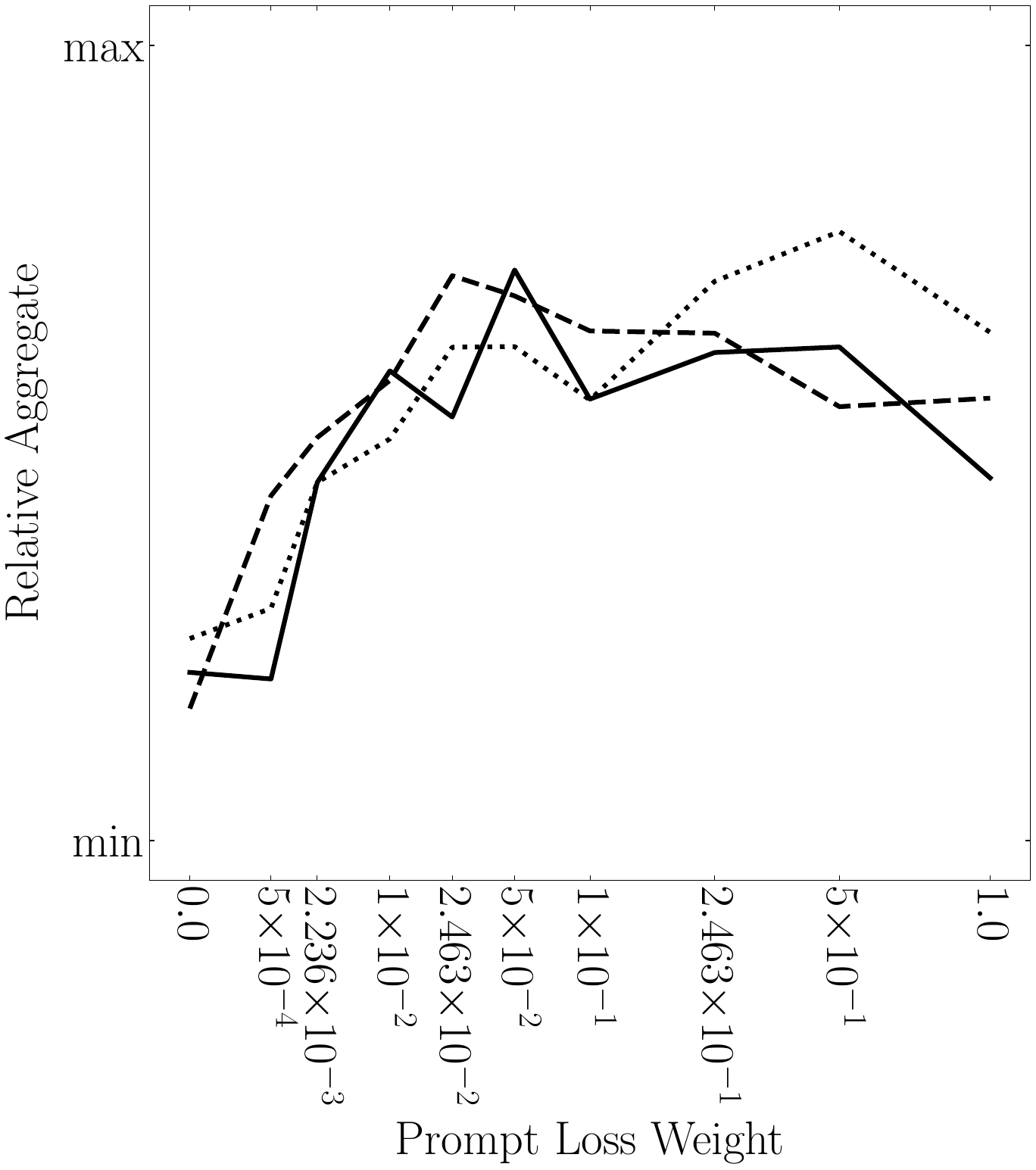}
        \label{fig:short-aug-short}
        \vspace{-0.6cm}
        \caption{Prompt-Inverted Datasets}
    \end{subfigure}
    \hspace{0.4cm}
    \begin{subfigure}[b]{.15\linewidth}
        \includegraphics[width=\linewidth]{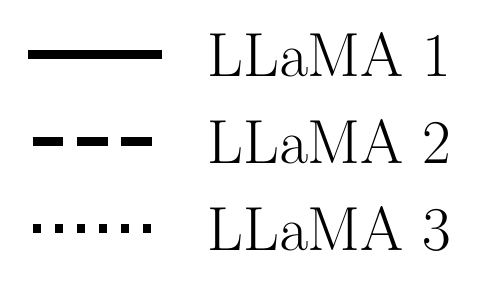}
        \vspace{3.3cm}
    \end{subfigure}
    \hfill
    \caption{
    Relative aggregate scores showing the effects of PLW for SIFT on alternative datasets.
    \textbf{(a)} UltraFeedbackCleaned and DatabricksDolly models.
    \textbf{(b)} UltraFeedbackShort and DatabricksDollyShort models.
    }
    \label{fig:short-aug}
\end{figure*}

We repeated our main training experiment using two additional datasets, a cleaned and binarized version of UltraFeedback (\citealp{cui2023ultrafeedback}; denoted UltraFeedbackCleaned) and databricks-dolly-15k (\citealp{DatabricksBlog2023DollyV2}; denoted DatabricksDolly).
We also trained on prompt-inverted versions of both datasets, denoted UltraFeedbackShort and DatabricksDollyShort, respectively, and expanded analysis to include LLaMA 3 8B \cite{llama3modelcard} as a PTLM.
See length statistics for these four datasets in table~\ref{tab:data} and Appendix~\ref{sec:app-aug} for additional visualizations.
Note that compared with the unmodified datasets used in the main experiment, both UltraFeedback and DatabricksDolly have significantly lower generation ratios of $R_g = 1.77$ and $R_g = 0.83$, respectively.
These datasets were chosen to demonstrate the effects of PLW when fine-tuning on data with relatively balanced generation ratios.

The combined relative aggregate scores for models trained on UltraFeedbackCleaned and DatabricksDolly and for models trained on the prompt-inverted UltraFeedbackShort and DatabricksDollyShort datasets can be seen in figure~\ref{fig:short-aug}.
Visual inspection suggests that PLW affected learning for both groups of datasets.
For the shortened variants, performance appears to have the same negative quadratic relationship with PLW as the AlpacaDataShort models did in the main experiment.
The relationship between PLW and performance for models fine-tuned on the unmodified datasets is weaker, but appears to be generally increasing.

The clear relationship in the shortened data variants shows that the results of the main experiment extend to additional datasets.
Furthermore, the weak relationship between PLW and performance of the unmodified data variants suggests that the effects of PLW are indeed dependent on the generation ratio.
See Appendix~\ref{sec:app-gam} for an approach to predicting optimal PLW based on the generation ratio of the SIFT dataset.

\section{Conclusion}
\label{sec:conclusion}

In this study, we explored the effects of prompt loss weight (PLW) on LLM supervised instruction fine-tuning (SIFT).

We found that PLW had a statistically significant effect on learning for our short-completion dataset, and proper tuning of PLW allowed short-completion-trained models to outperform long-completion-trained models on short-generation benchmarks.
We showed that the causal mechanism was due to a balance between two different regularizing effects and not due to increased training stability as is commonly attributed.
We showed that the measured relationship extends to additional SIFT datasets and that the effects could not be sufficiently emulated with other regularizers.

Based on the above conclusions, we assert the following two points.
\begin{enumerate}
    \item Since models fine-tuned on short-completion datasets and with properly tuned PLWs outperformed all other models on short-generation benchmark tasks, we conclude that PLW is critical for effectively fine-tuning for downstream short-generation tasks.
    \item Given the importance of PLW and given that many SIFT datasets and almost all natural language understanding (NLU) datasets are short-completion datasets, we warn SIFT API providers about the need for a PLW parameter to adequately cover a full range of modeling applications.
\end{enumerate}

\newpage

\section*{Limitations}

\begin{enumerate}
    \item We analyzed prompt loss weighting (PLW) for instruction fine-tuning LLMs.
    We characterized seven fine-tuning datasets by their relative completion-prompt length ratios and reported on the effect of PLW when training on each dataset.
    It would be helpful to extend this research to a wider range of datasets to increase the strength of our conclusions and create more complete guidelines for prompt loss weighting.
    \item Since we used pre-trained models and no layers were freshly initialized, there was little variance in initial experiments.
    We therefore limited runs to a single seed of $42$.
    \item Suggested values for PLW from section~\ref{sec:app-gam} are based on the included experiments.
    Best PLW values when fine-tuning different models or using different datasets or training regimes may vary from the relationships shown here, though we are still confident that performance will not vary significantly by PLW for long-completion data.
    \item The focus of our research was on how PLW affected fine-tuning based on the completion-prompt \emph{ratio} of the training dataset. However, the absolute length and size of the dataset will likely play a role in learning dynamics. It would be good to include that perspective in future research on token loss weights.
    \item LLM-as-evaluator approaches like PandaLM and AlpacaEval are still relatively new, and these approaches are being actively developed.
    We chose to use Mixtral 8x7B as an auto-evaluator for AlpacaEval 1 due to budget limitations.
    While we cite high human evaluation correlation with Mixtral and justify this decision in section~\ref{sec:eval}, using the default auto-evaluator would be beneficial for better comparison with other research.
    \item While we define short- and long-completion data as have a completion-prompt ratio $R_g$ lower and greater than $1$, respectively, we do not provide justification for using $1$ as the threshold. Choosing a meaningful reference would be helpful to future research.
\end{enumerate}

\section*{Ethical Considerations}

In this paper, we presented an analysis of the prompt loss weight hyperparameter for supervised instruction fine-tuning.
We did not rely on human evaluators, and at no point in our research did we expose anyone to risk of harm.

We acknowledge that standard deep learning training methods have a high carbon footprint, and we performed over 200 fine-tuning training runs.
Model outputs cannot be predicted in advance, and, while we release our model weights in the spirit of transparency and collaboration, models may hallucinate or produce offensive output.
Additionally, our shortened datasets were generated from publicly released data, and we did not perform additional content filtering.
A warning about both of these issues will be released along with the models and datasets.

\section*{Acknowledgements}

This work was supported by the Technology Innovation Program funded by the Korean Ministry of Trade, Energy, and Industry (MOTIE, Korea) (No. 20014406, ``Development of interactive sign language interpretation service based on artificial intelligence for the hearing impaired'') and the Artificial intelligence industrial convergence cluster development project funded by the Ministry of Science and ICT (MSIT, Korea) \& Gwangju Metropolitan City (No. BA00000797, LLM-based sign language translation for weather forecasts).

\bibliography{paper}

\clearpage

\appendix

\onecolumn

\startcontents[appendices]
\printcontents[appendices]{l}{1}{\section*{Appendices}\setcounter{tocdepth}{2}}

\clearpage
\twocolumn
\section{Prompt Inversion}\label{sec:app-short}

\begin{figure*}[!tb]
    \centering
    \includegraphics[width=\linewidth]{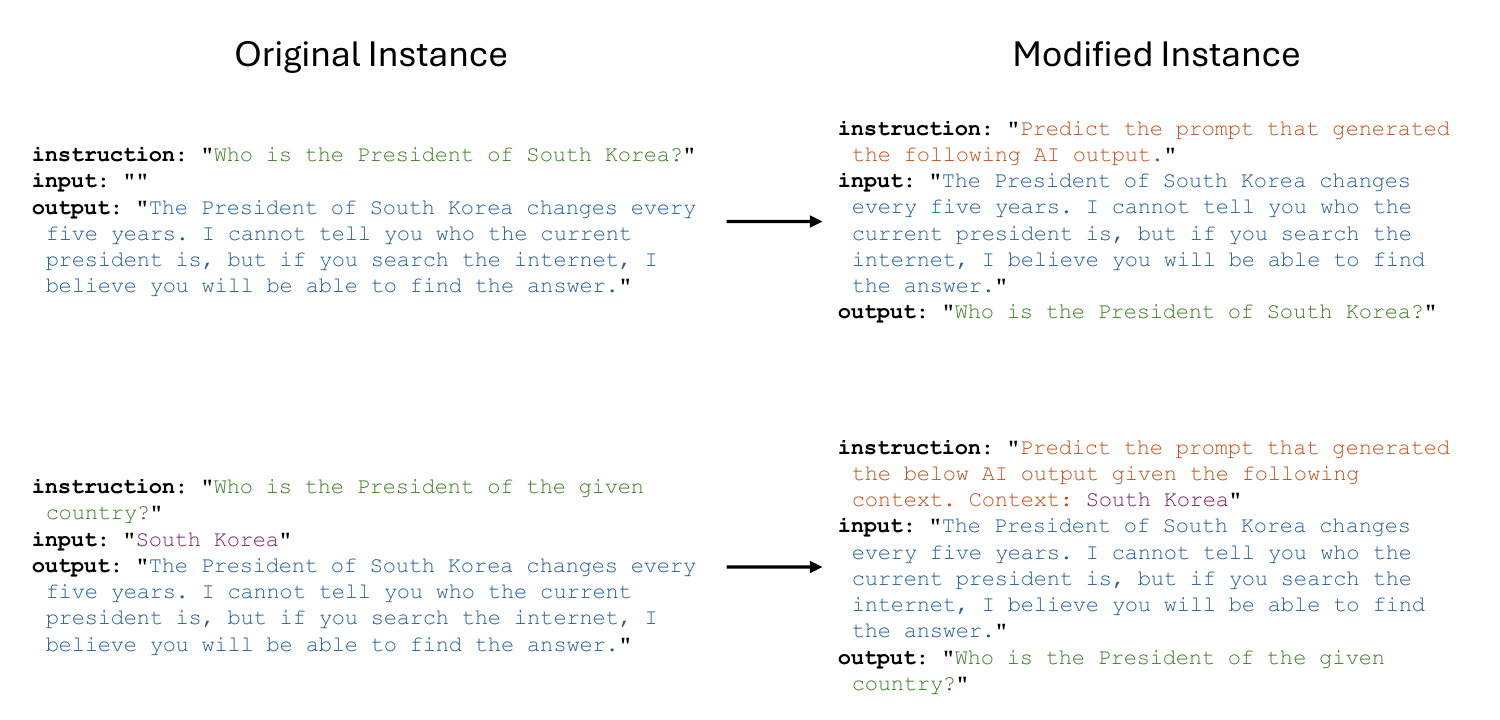}
    \caption{
    Examples of modifying prompt-completion ratios using prompt inversion, best viewed in color.
    To prompt-invert instances, we re-frame the prompt-completion task as an original-prompt-prediction task.
    I.e., we teach the model to predict the original instruction given an example completion and optional input.
    In the first example above, prompt inversion changes the instance's word-based completion-prompt ratio $R_g$ from $34/(7+0)=4.857$ to $7/(9+34)=0.163$.
    }
    \label{fig:app-short}
\end{figure*}

In order to experiment with short-completion fine-tuning datasets, we propose using \emph{prompt inversion} to rotate the instruction, input, and output fields of instances.
Prompt inversion modifies an instance to use the original instruction as the new completion text and the original output as the new input text.
The model is then given the following instruction:\\
``\texttt{Predict the prompt that generated the following AI output.}''\\
if the input field is empty and\\
``\texttt{Predict the prompt that generated the below AI output given the following context. Context: <original-input>}''\\
if there is an input field.
See figure~\ref{fig:app-short} for a visualization of this process.

Synthesis of the original input and output texts to predict the original instruction requires both language understanding and reasoning, and prompt-inverted instances should be seen as natural language understanding (NLU) tasks.

To generate short versions of the instruction datasets used in our experiments, we used prompt-inversion to modify every instance with a completion-prompt length ratio $R_g > 1$, based on the tokenized lengths of each field (where ``prompt'' is the concatenation of the instruction and input fields as explained in section~\ref{sec:def}).
Thus, given any long-completion dataset, a textually-similar short-completion dataset can be generated and used for comparison.
Note that unless all instances have a generation ratio $R_g > 1$, the resulting dataset will contain a mixture of unmodified instances and prompt-inverted instances.

\clearpage
\section{Main Experiment Benchmarks}
\label{sec:app-taskperf}

This section presents additional qualitative analysis of benchmark performance and score visualizations for each benchmark.

For both the simple aggregate and the relative aggregate, models trained on AlpacaDataShort showed a visual relationship with \texttt{w\textsubscript{p}}.
Based on this visual relationship, we divide benchmarks into three groups.

\begin{figure*}[tb!]
    \begin{subfigure}[b]{.32\linewidth}
        \includegraphics[width=\linewidth]{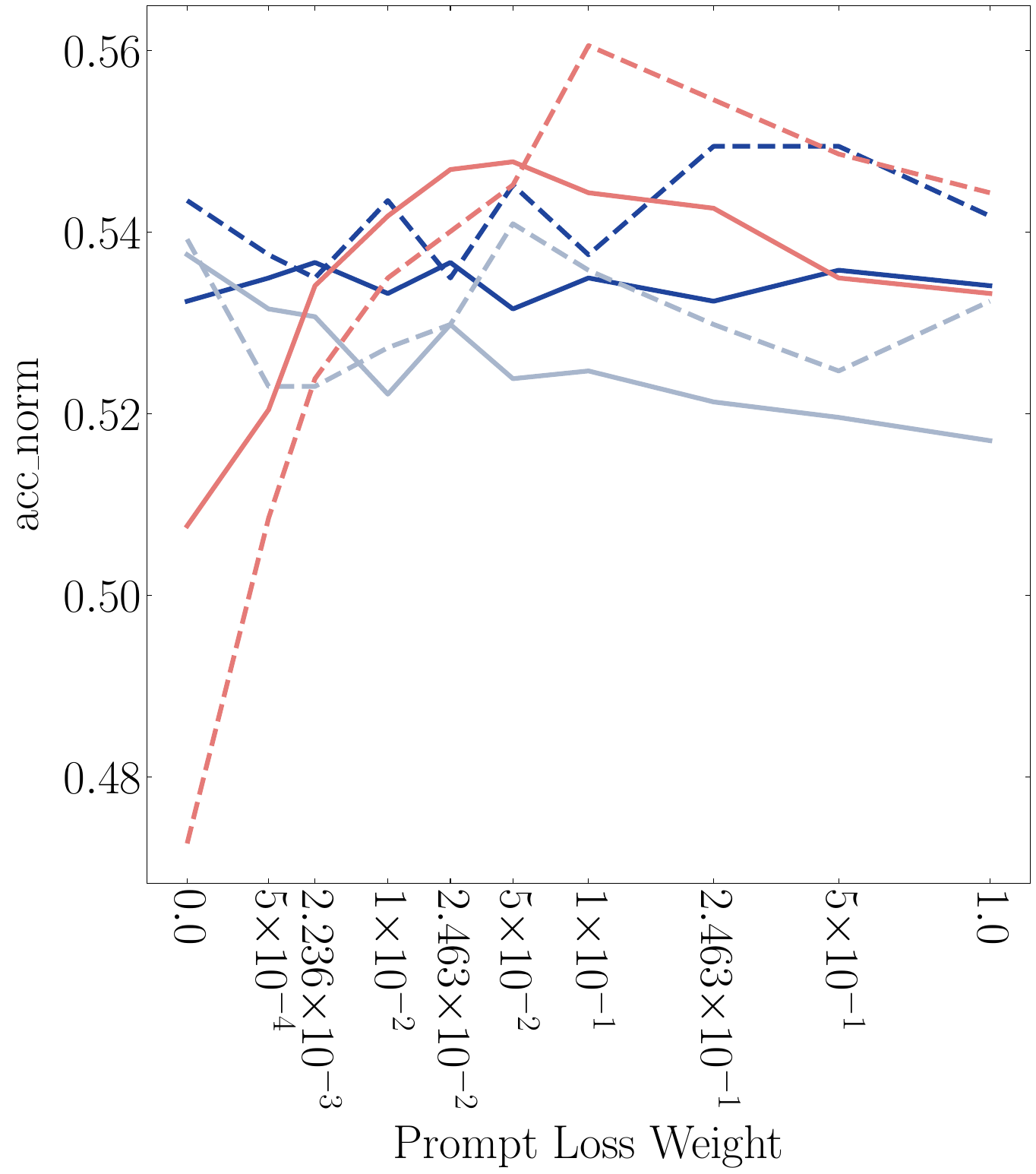}
        \vspace{-0.6cm}
        \caption{ARC Challenge}
    \end{subfigure}
    \hfill
    \begin{subfigure}[b]{.328\linewidth}
        \includegraphics[width=\linewidth]{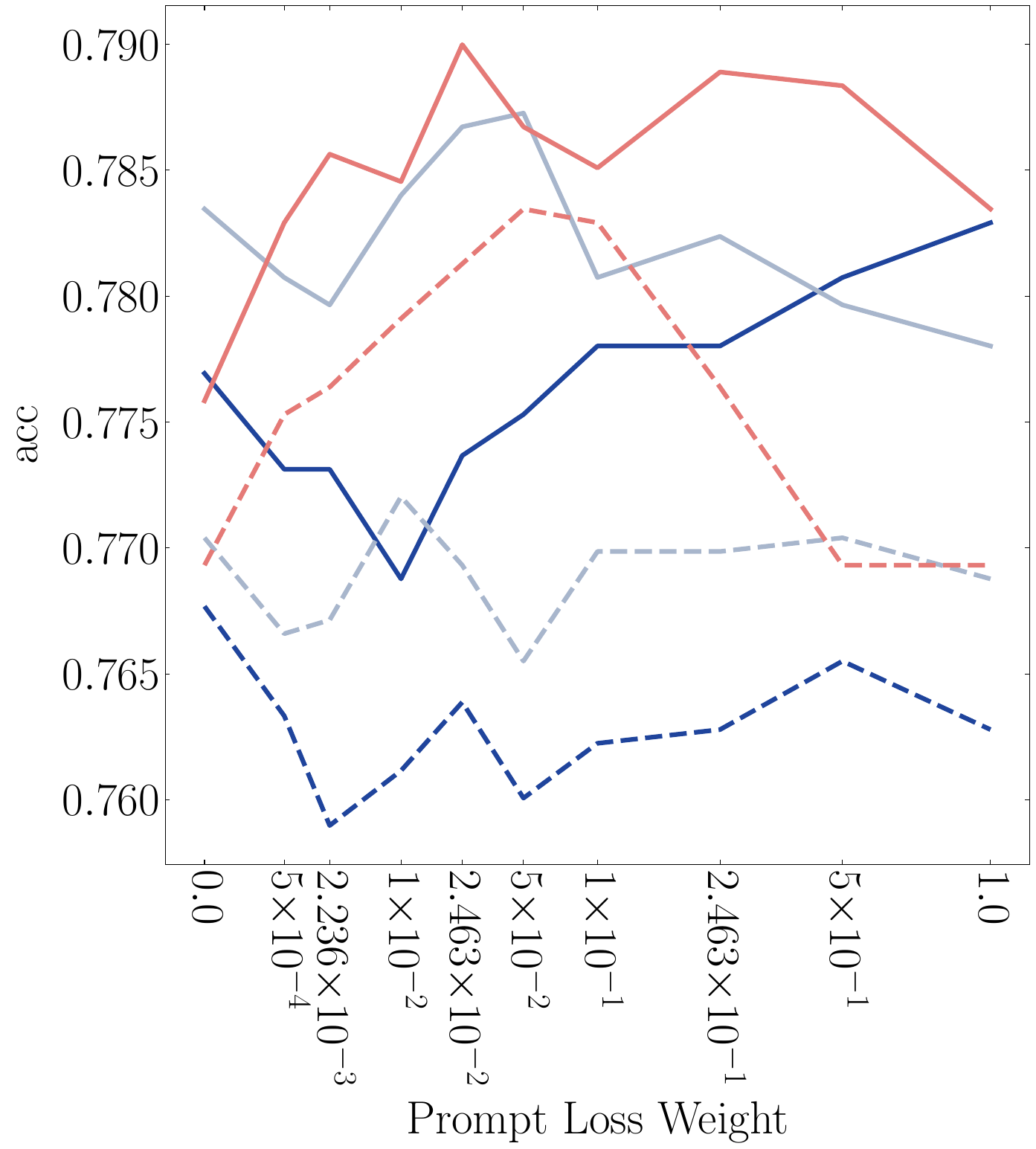}
        \vspace{-0.6cm}
        \caption{PIQA}
    \end{subfigure}
    \hfill
    \begin{subfigure}[b]{.32\linewidth}
        \includegraphics[width=\linewidth]{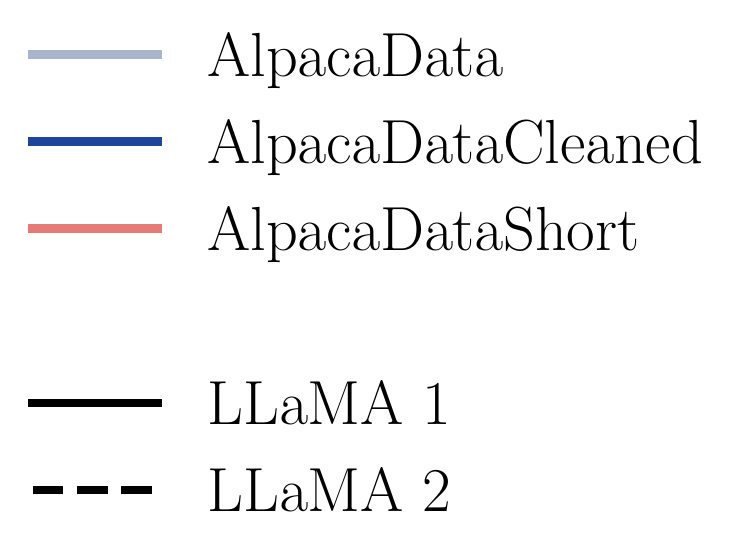}
        \vspace{1.5cm}
    \end{subfigure} \\ \\
    \begin{subfigure}[b]{.32\linewidth}
        \includegraphics[width=\linewidth]{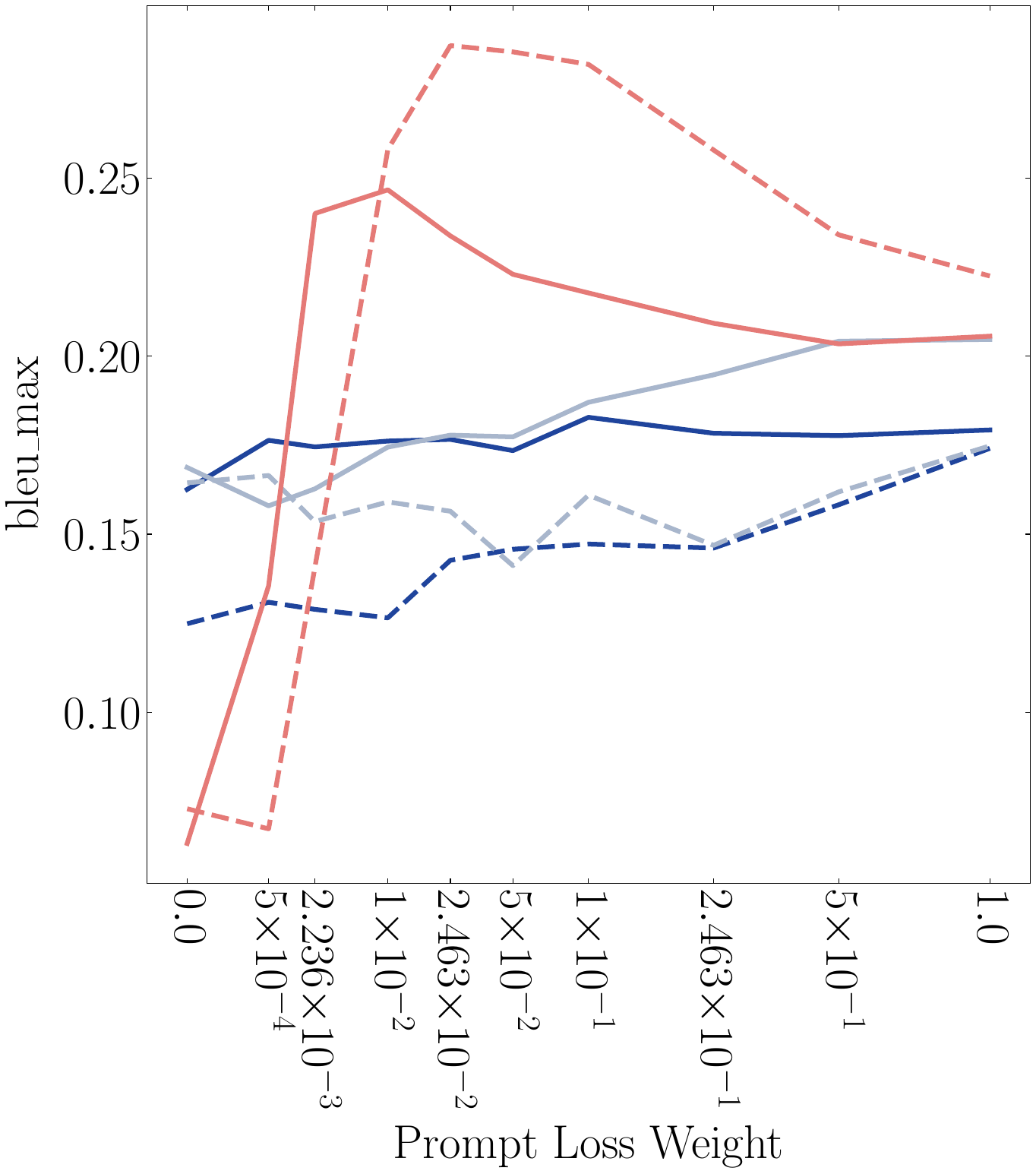}
        \vspace{-0.6cm}
        \caption{TruthfulQA-Gen}
    \end{subfigure}
    \hspace{0.25cm}
    \begin{subfigure}[b]{.32\linewidth}
        \includegraphics[width=\linewidth]{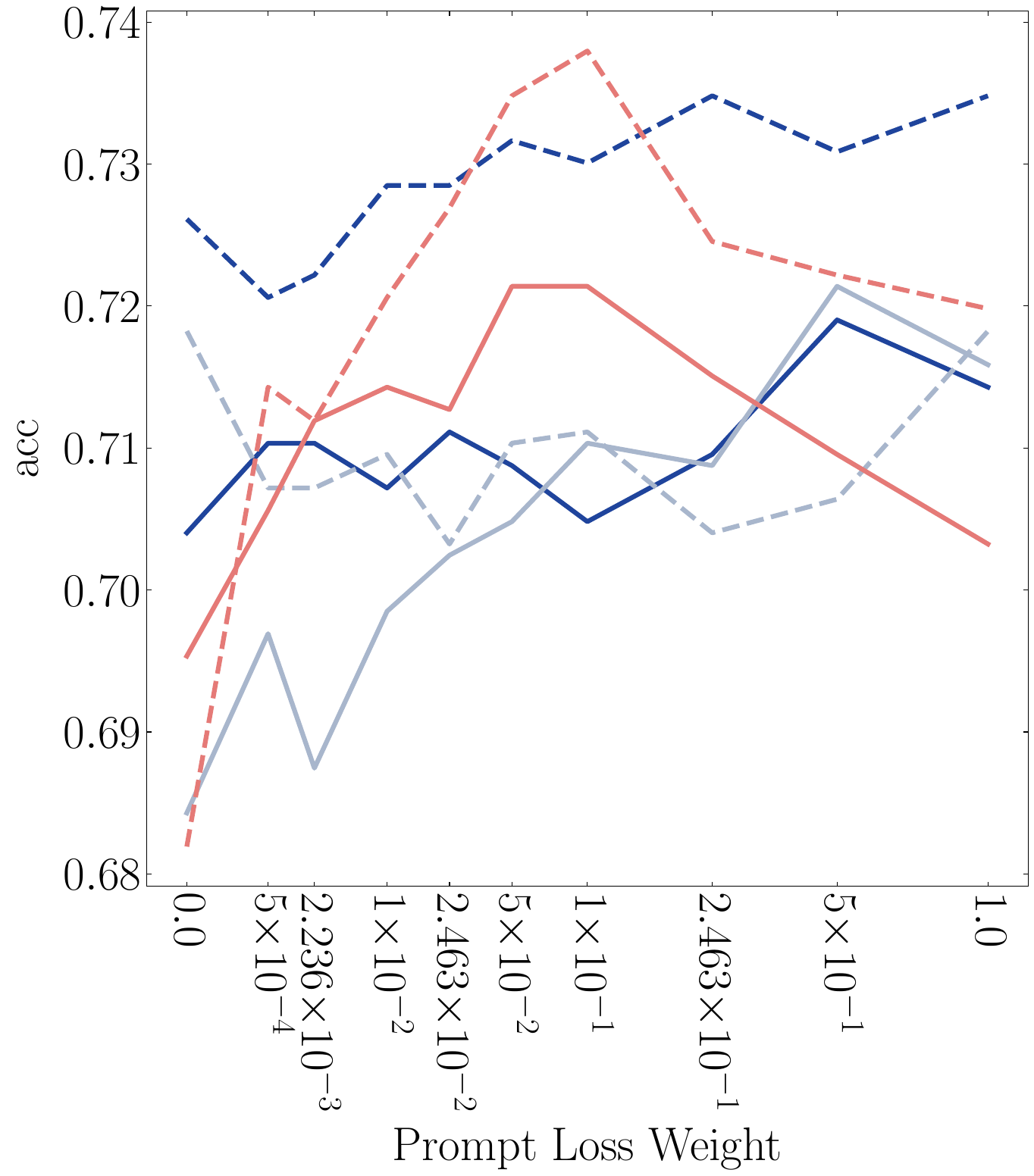}
        \vspace{-0.6cm}
        \caption{WinoGrande}
    \end{subfigure}
    \caption{
    Group I benchmark performance.
    Note the negative quadratic relationship with transformed PLW.
    }
    \label{fig:task-quad}
\end{figure*}

The first group showed a negative quadratic relationship with \texttt{w\textsubscript{p}}, with performance \textbf{exceeding} that of AlpacaDataCleaned models.
This group consists of ARC Challenge, PIQA, TruthfulQA-Gen, and WinoGrande benchmarks, and optimal PLW values for these four benchmarks vary from PLW $=0.01$ to PLW $=0.1$.
See figure~\ref{fig:task-quad} for individual benchmark visualizations.

\begin{figure*}[tb!]
    \centering
    \begin{subfigure}[b]{.32\linewidth}
        \includegraphics[width=\linewidth]{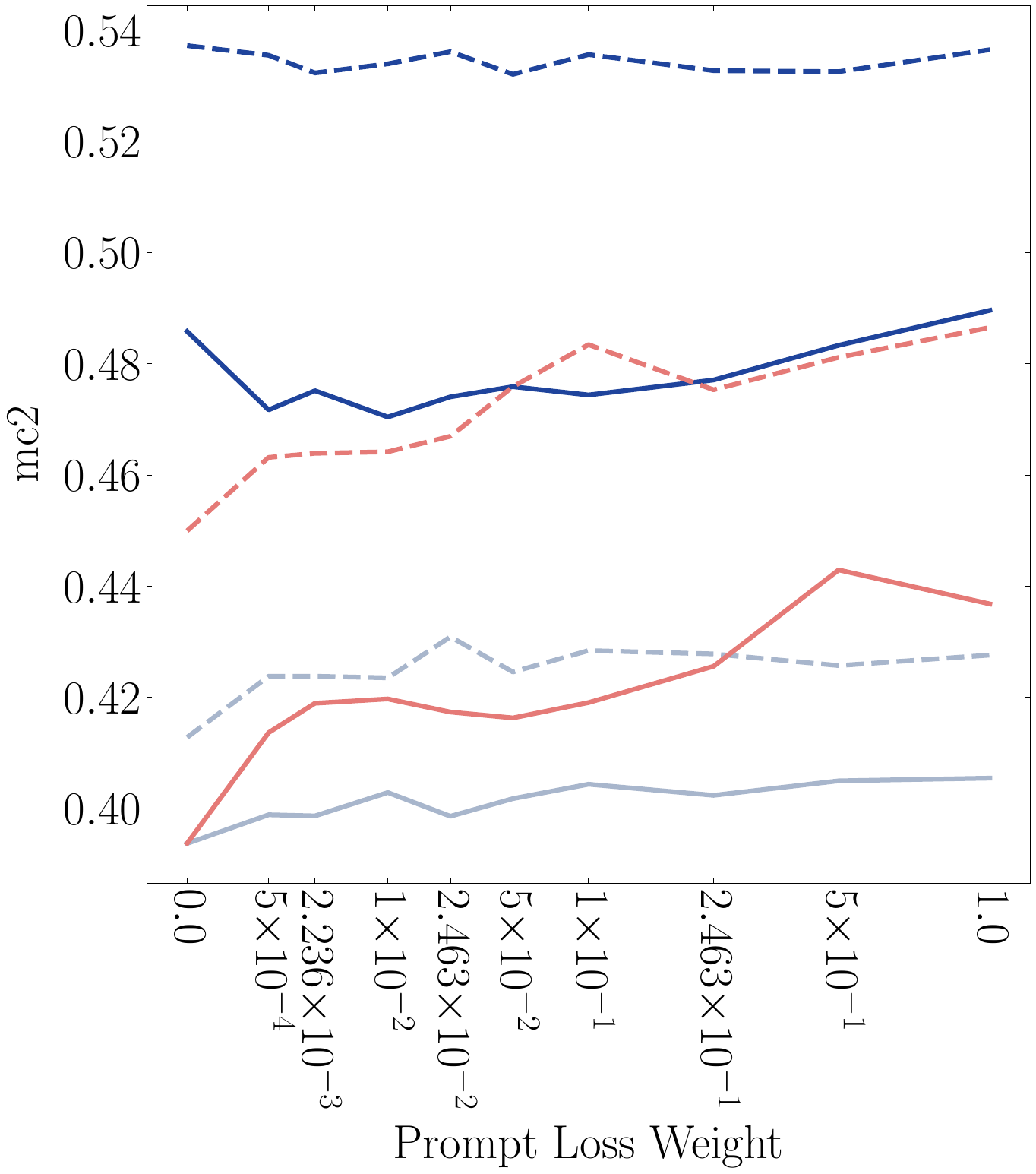} 
        \vspace{-0.6cm}
        \caption{TruthfulQA-MC2}
    \end{subfigure} 
    \hspace{0.7cm}
    \begin{subfigure}[b]{.32\linewidth}
        \includegraphics[width=\linewidth]{images/legend_space.pdf}
        \vspace{1.5cm}
    \end{subfigure} \\
    \vspace{5mm}
    \begin{subfigure}[b]{.32\linewidth}
        \includegraphics[width=\linewidth]{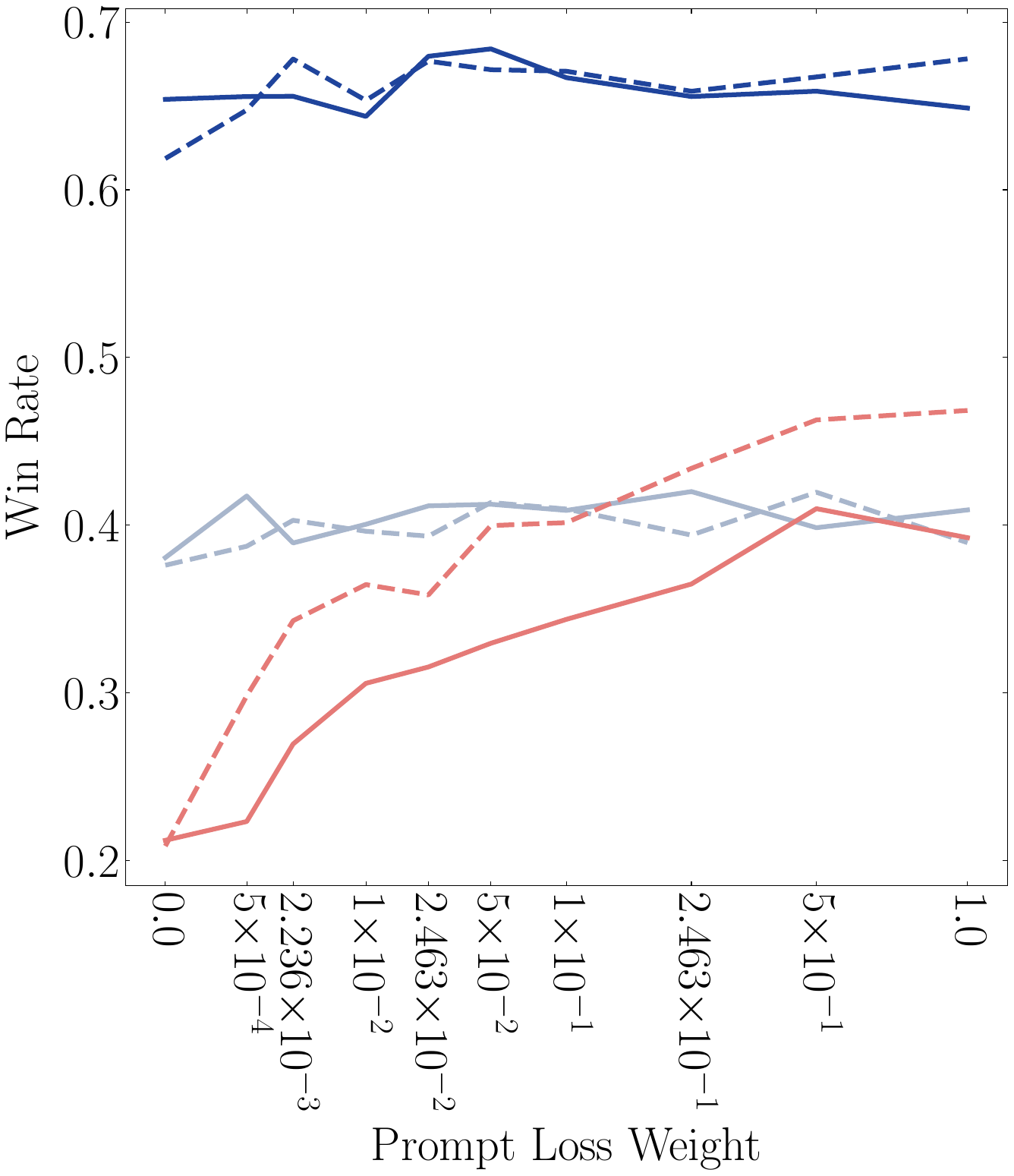}
        \vspace{-0.6cm}
        \caption{Alpaca Eval (AE) v1}
    \end{subfigure} 
    \hspace{0.7cm}
    \begin{subfigure}[b]{.32\linewidth}
        \includegraphics[width=\linewidth]{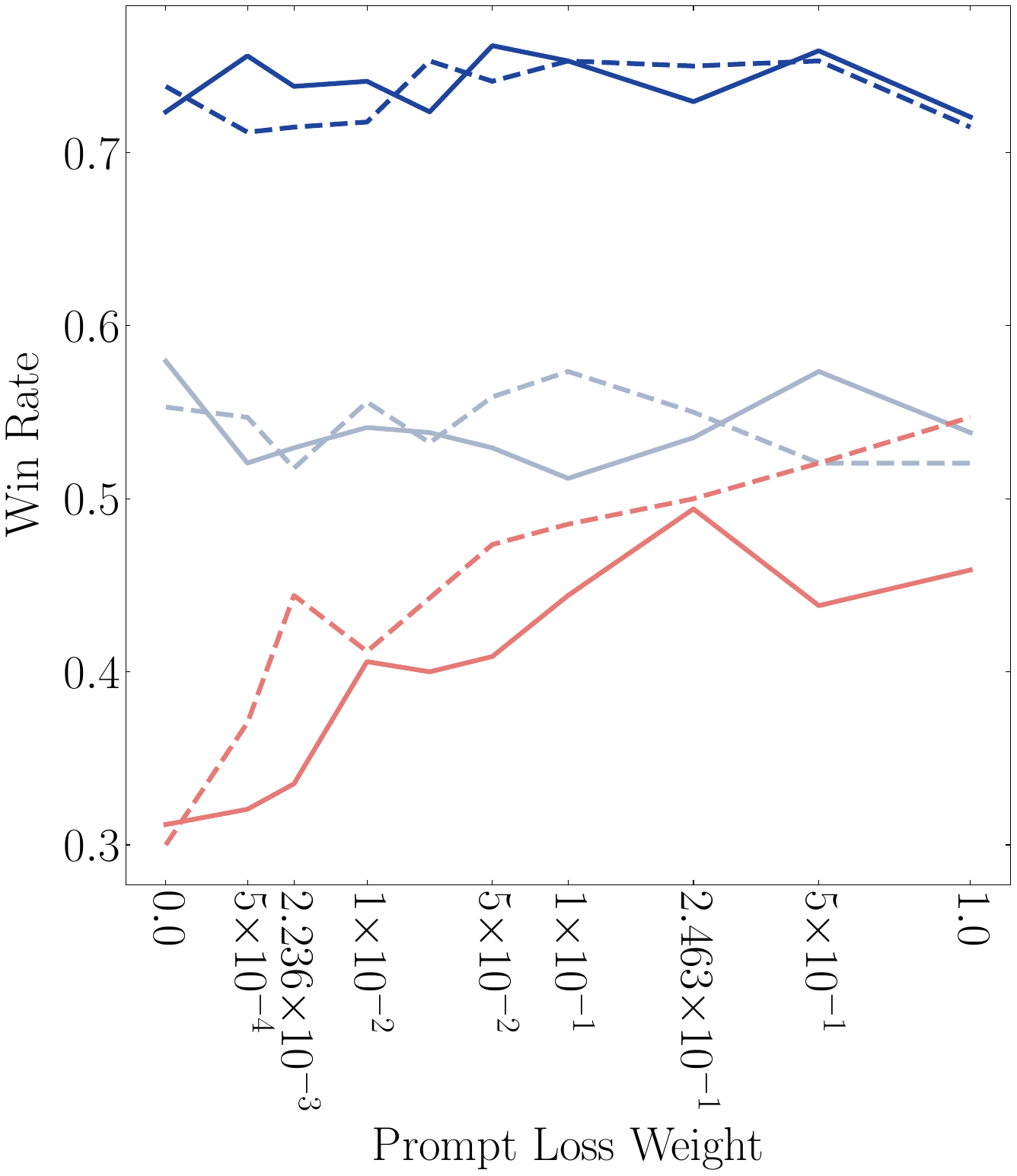}
        \vspace{-0.6cm}
        \caption{PandaLM}
    \end{subfigure} 
    \caption{
    Group II benchmarks showed increasing performance with PLW.
    }
    \label{fig:pos}
\end{figure*}

The second group of benchmarks showed steadily increasing performance as \texttt{w\textsubscript{p}} increased, before leveling off to maximum values near \texttt{w\textsubscript{p}}$=1$.
This group is TruthfulQA-MC2, AlpacaEval 1, and PandaLM.
It is surprising that TruthfulQA-MC2 shows a relationshp more similar to the long-generation benchmarks and TruthfulQA-Gen resembles the other multi-choice benchmarks.
See figure~\ref{fig:pos} for individual benchmark visualizations.

Interestingly, on the seven benchmarks from groups I and II, $\texttt{w\textsubscript{p}}>0$ was almost always better than $\texttt{w\textsubscript{p}}=0$ (i.e., complete masking led to the worst performance) for AlpacaDataShort models.

\begin{figure*}[tb!]
    \centering
    \begin{subfigure}[b]{.325\linewidth}
        \includegraphics[width=\linewidth]{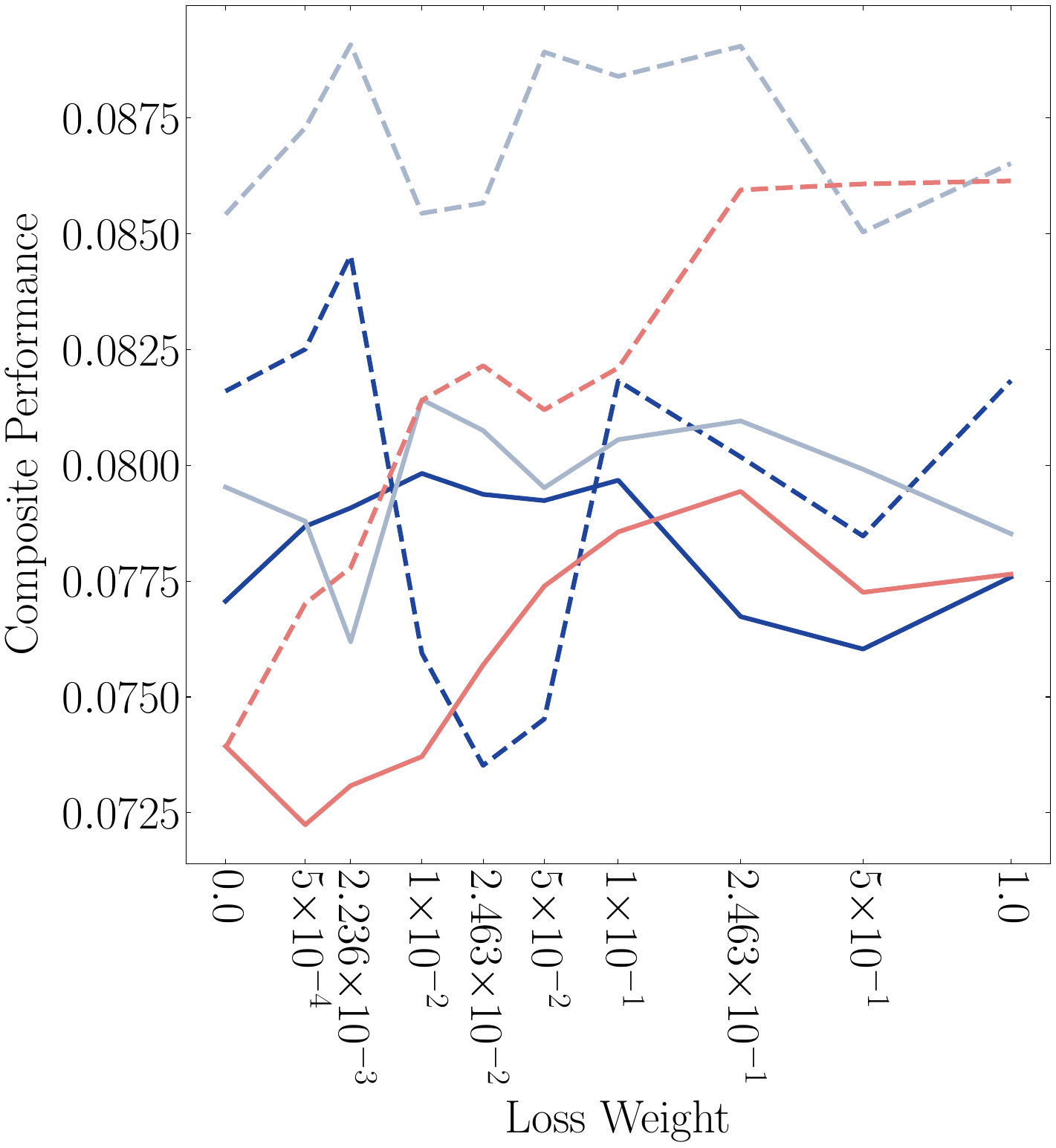}
        \vspace{-0.6cm}
        \caption{Translation From English}
        \label{fig:fromen}
    \end{subfigure} \hfill
    \begin{subfigure}[b]{.32\linewidth}
        \includegraphics[width=\linewidth]{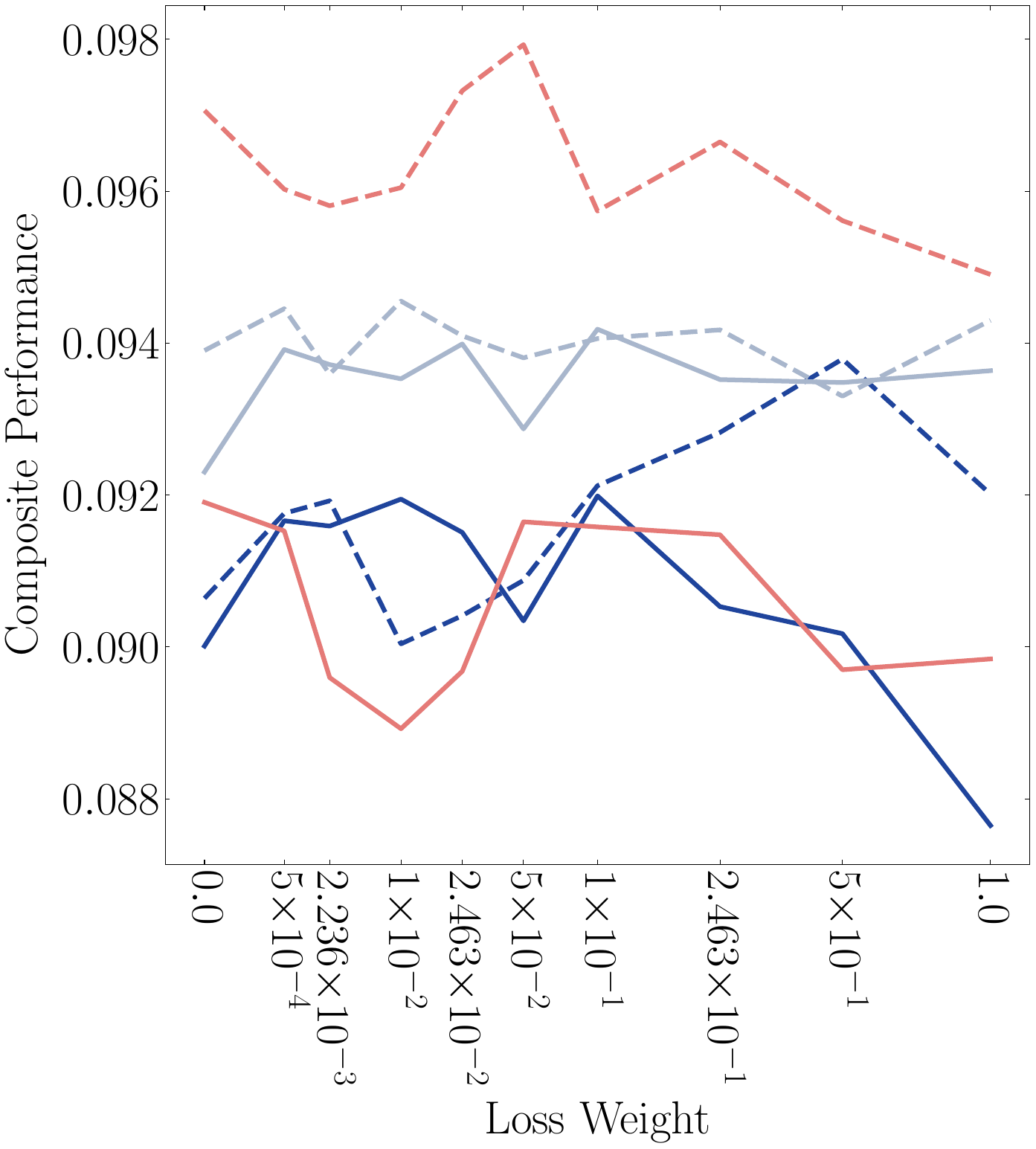}
        \vspace{-0.6cm}
        \caption{Translation To English}
        \label{fig:toen}
    \end{subfigure} \hfill
    \begin{subfigure}[b]{.32\linewidth}
        \includegraphics[width=\linewidth]{images/legend_space.pdf}
        \vspace{1.5cm}
    \end{subfigure} \\
    \vspace{5mm}
    \begin{subfigure}[b]{.325\linewidth}
        \includegraphics[width=\linewidth]{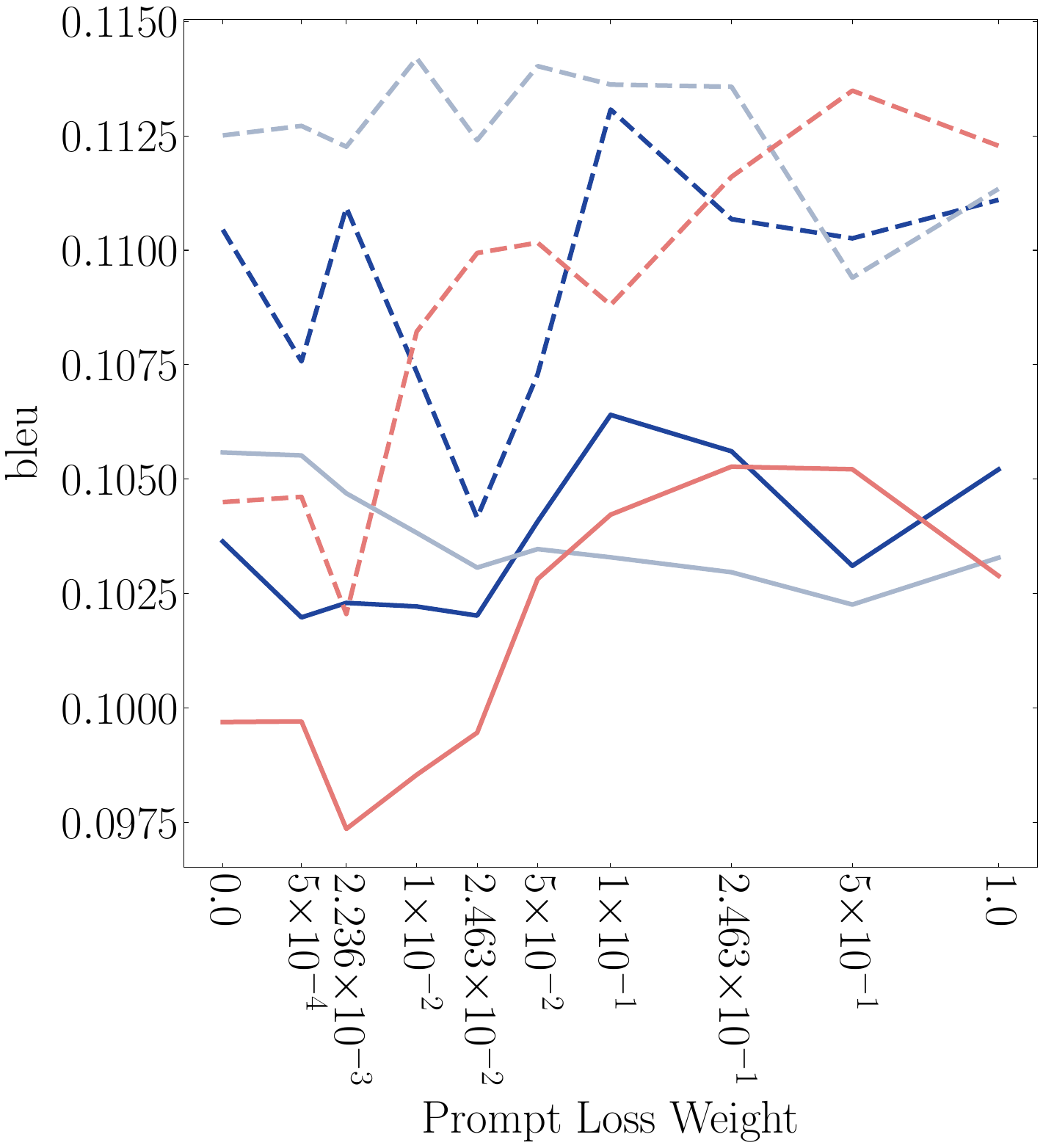}
        \vspace{-0.6cm}
        \caption{En$\rightarrow$Fr Translation}
    \end{subfigure} \hfill
    \begin{subfigure}[b]{.32\linewidth}
        \includegraphics[width=\linewidth]{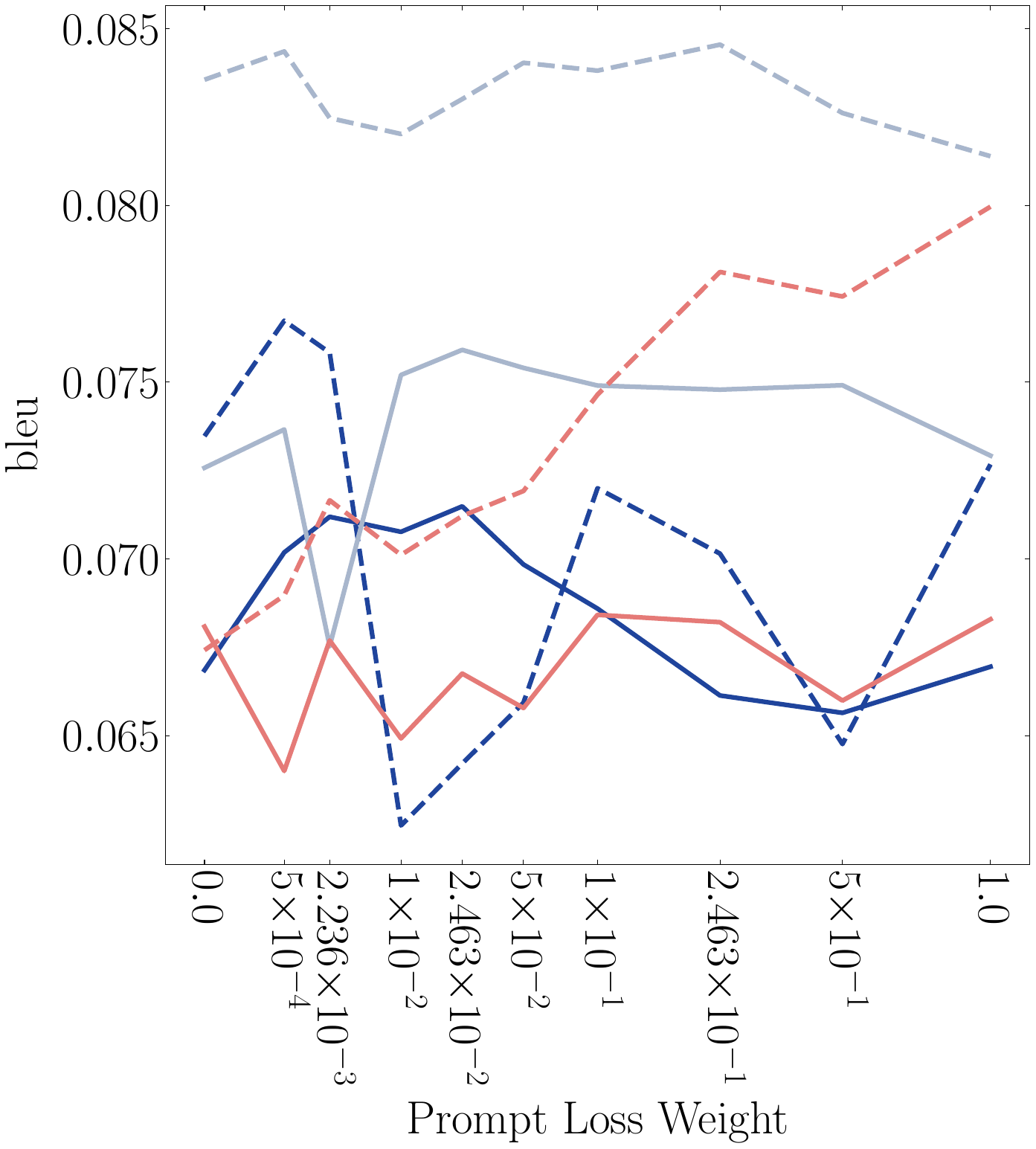}
        \vspace{-0.6cm}
        \caption{En$\rightarrow$De Translation}
    \end{subfigure} \hfill
    \begin{subfigure}[b]{.32\linewidth}
        \includegraphics[width=\linewidth]{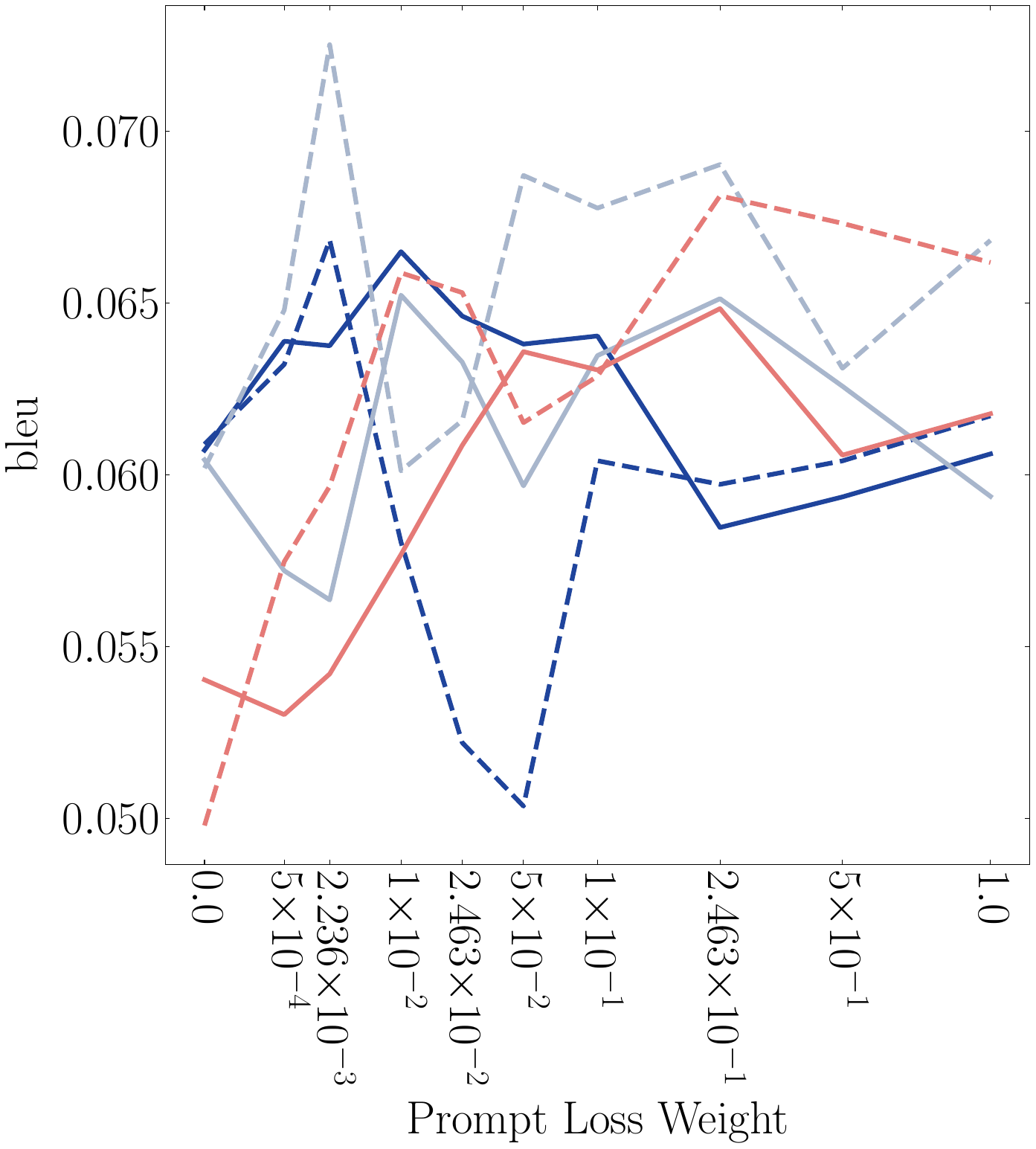}
        \vspace{-0.6cm}
        \caption{En$\rightarrow$Ro Translation}
    \end{subfigure} \\
    \vspace{5mm}
    \begin{subfigure}[b]{.32\linewidth}
        \includegraphics[width=\linewidth]{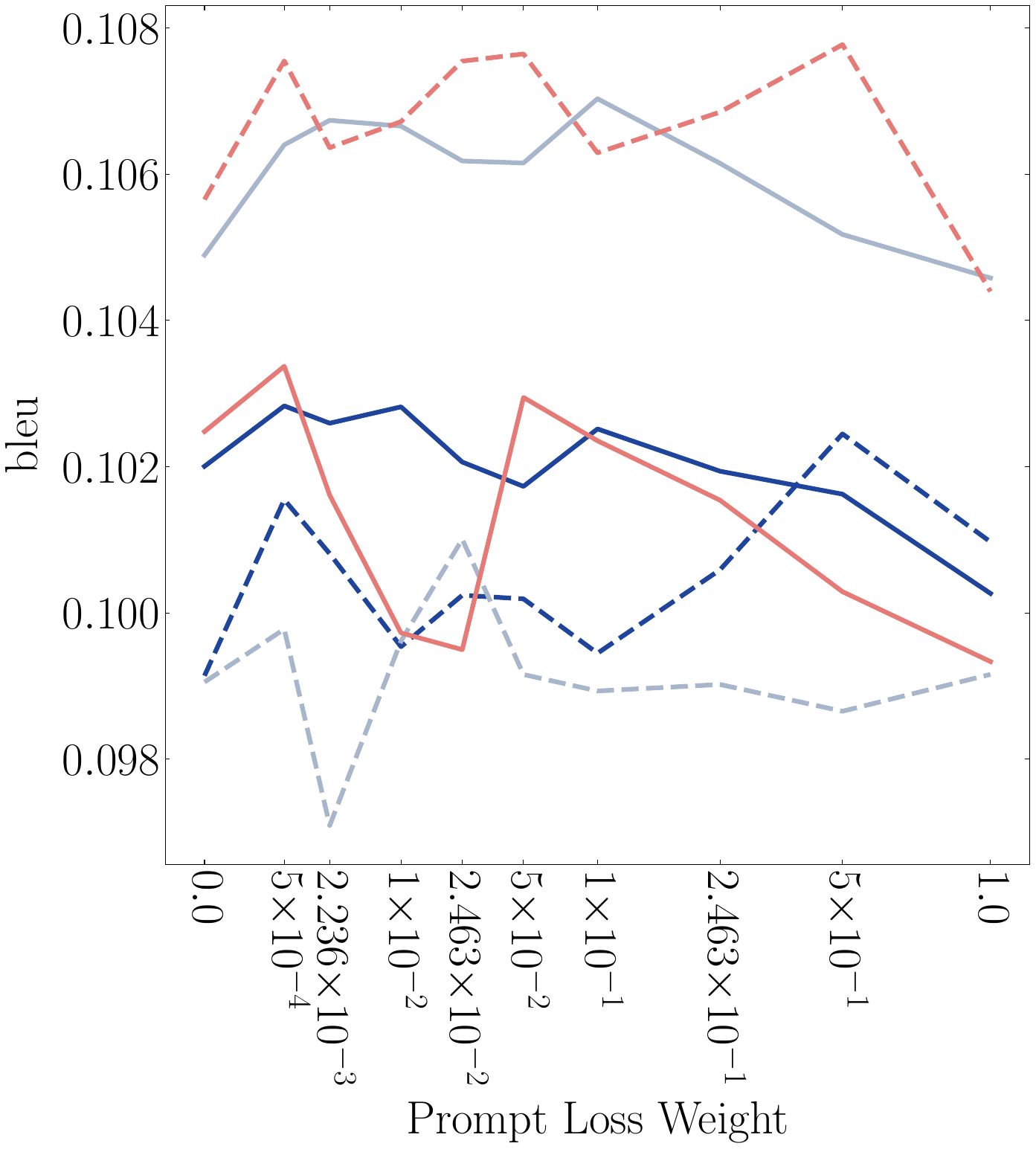}
        \vspace{-0.6cm}
        \caption{Fr$\rightarrow$En Translation}
    \end{subfigure} \hfill
    \begin{subfigure}[b]{.32\linewidth}
        \includegraphics[width=\linewidth]{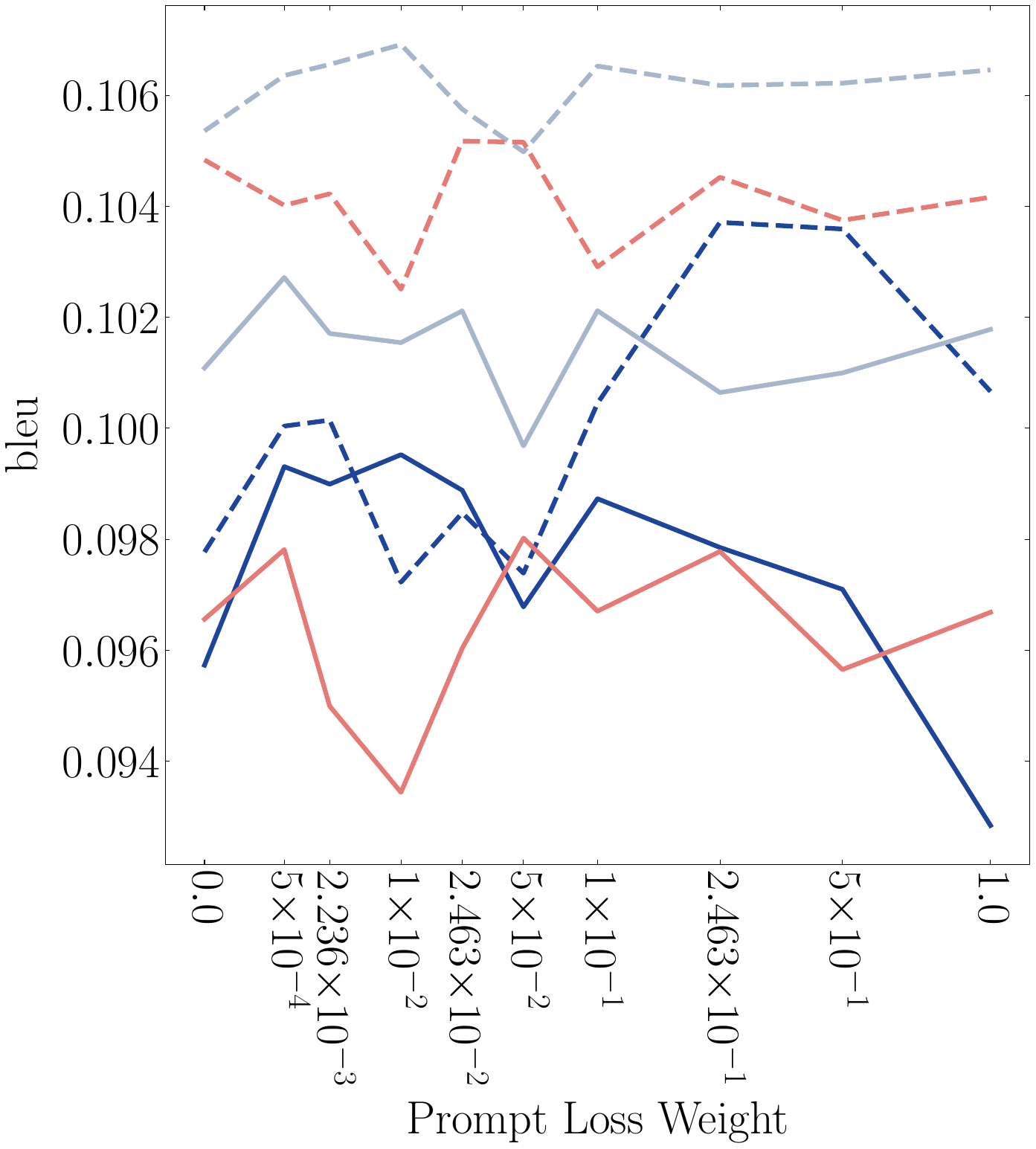}
        \vspace{-0.6cm}
        \caption{De$\rightarrow$En Translation}
    \end{subfigure} \hfill
    \begin{subfigure}[b]{.32\linewidth}
        \includegraphics[width=\linewidth]{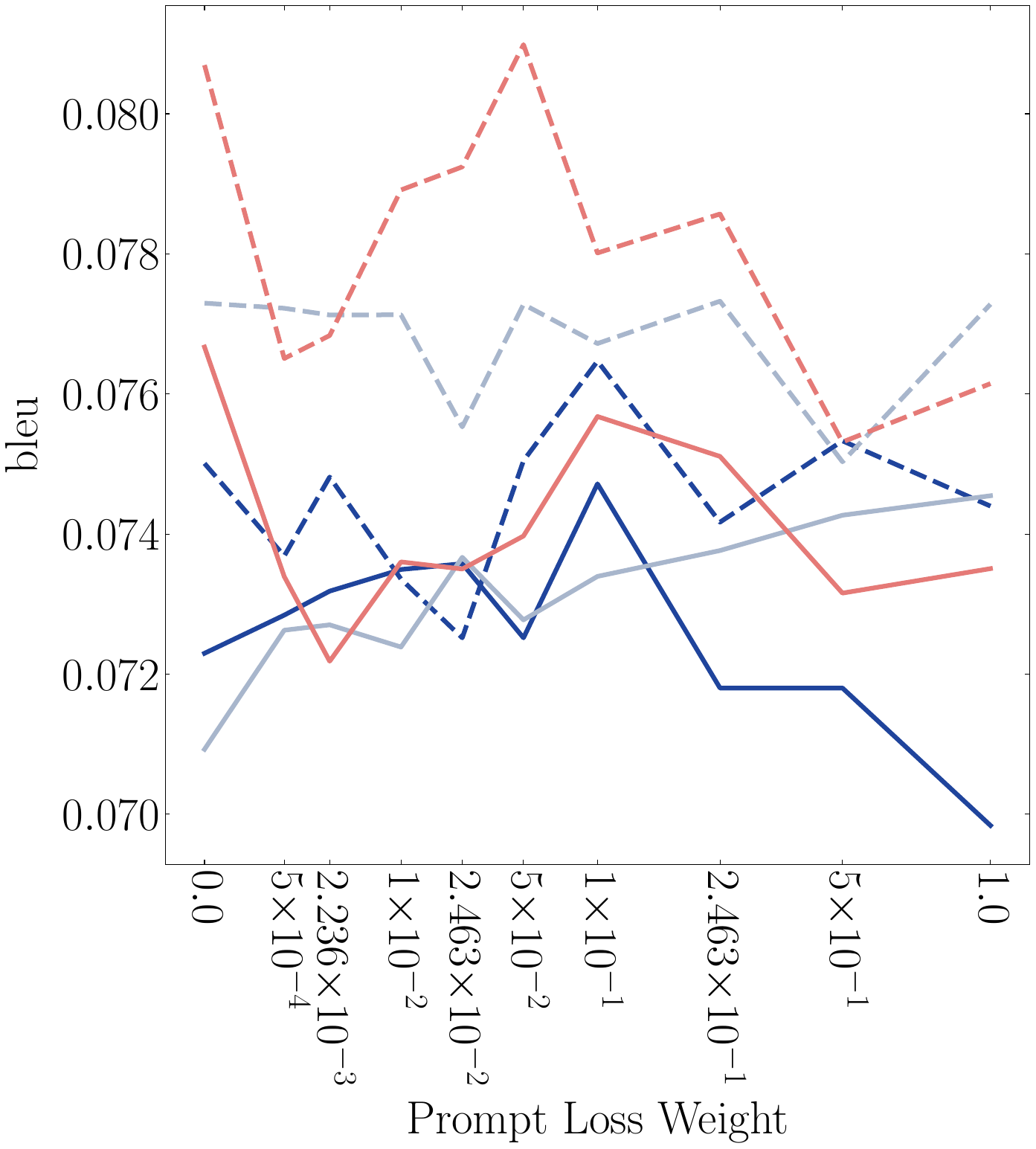}
        \vspace{-0.6cm}
        \caption{Ro$\rightarrow$En Translation}
    \end{subfigure}
    \caption{
    Group III benchmarks showed little relationship between performance and PLW.
    }
    \label{fig:noise}
\end{figure*}

The third group consists of the six translation benchmarks and showed unclear correlation between performance and \texttt{w\textsubscript{p}}.
Though aggregating benchmarks into ``to English'' and ``from English'' subgroups creates visualizations suggestive of a relationship, benchmarks from this group showed relatively more noise than the other benchmarks.
To reduce score noise, translation benchmarks were evaluated on the combined validation and test data splits, but there was still significant noise in the results.
See figure~\ref{fig:noise} for individual benchmark visualizations.

The performance difference across different \texttt{w\textsubscript{p}} values for the two long-generation benchmarks was around twenty percentage points, in stark contrast to the less than two percentage point change for short-generation and multiple choice benchmarks.
This suggests that PLW plays an important role in the ability to generate high quality text, and the optimal PLW for short-generation and long-generation benchmarks is clearly different.
Also note that performance of LLaMA 2 models was in general higher than that of LLaMA 1 models and performance of AlpacaDataCleaned models were higher than that of AlpacaData models, validating the improvements of LLaMA 2 and AlpacaDataCleaned over their predecessors.

\clearpage

\section{Visualizations for Supplemental Experiments}

This appendix contains visualizations and additional details about the two supplemental experiments from section~\ref{sec:supp}.

\begin{figure*}[tb!]
    \begin{subfigure}[b]{.3\linewidth}
        \includegraphics[width=\linewidth]{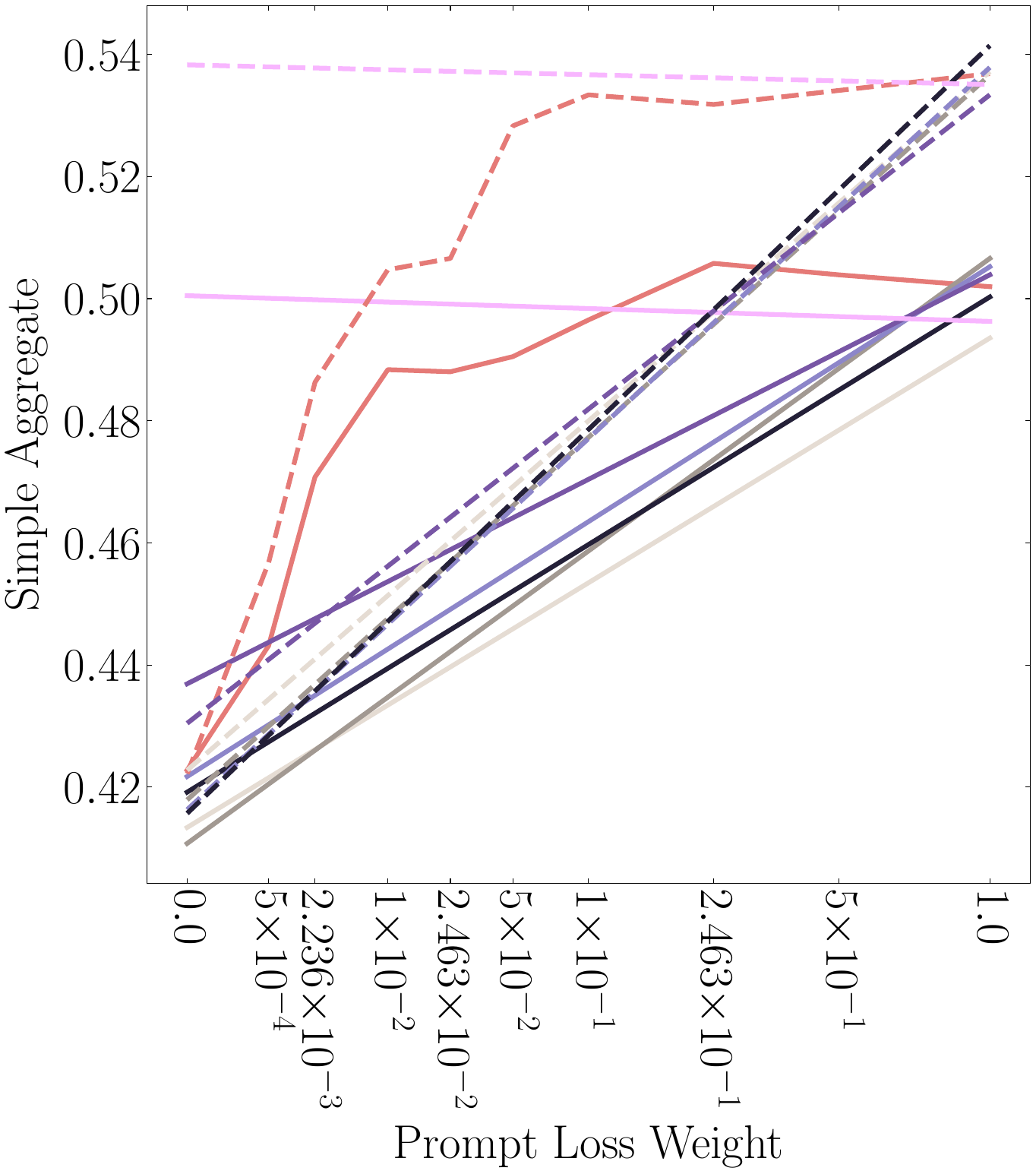}
        \vspace{-0.6cm}
        \caption{Simple Aggregate}
    \end{subfigure}
    \hspace{0.01\linewidth}
    \begin{subfigure}[b]{.3\linewidth}
        \includegraphics[width=\linewidth]{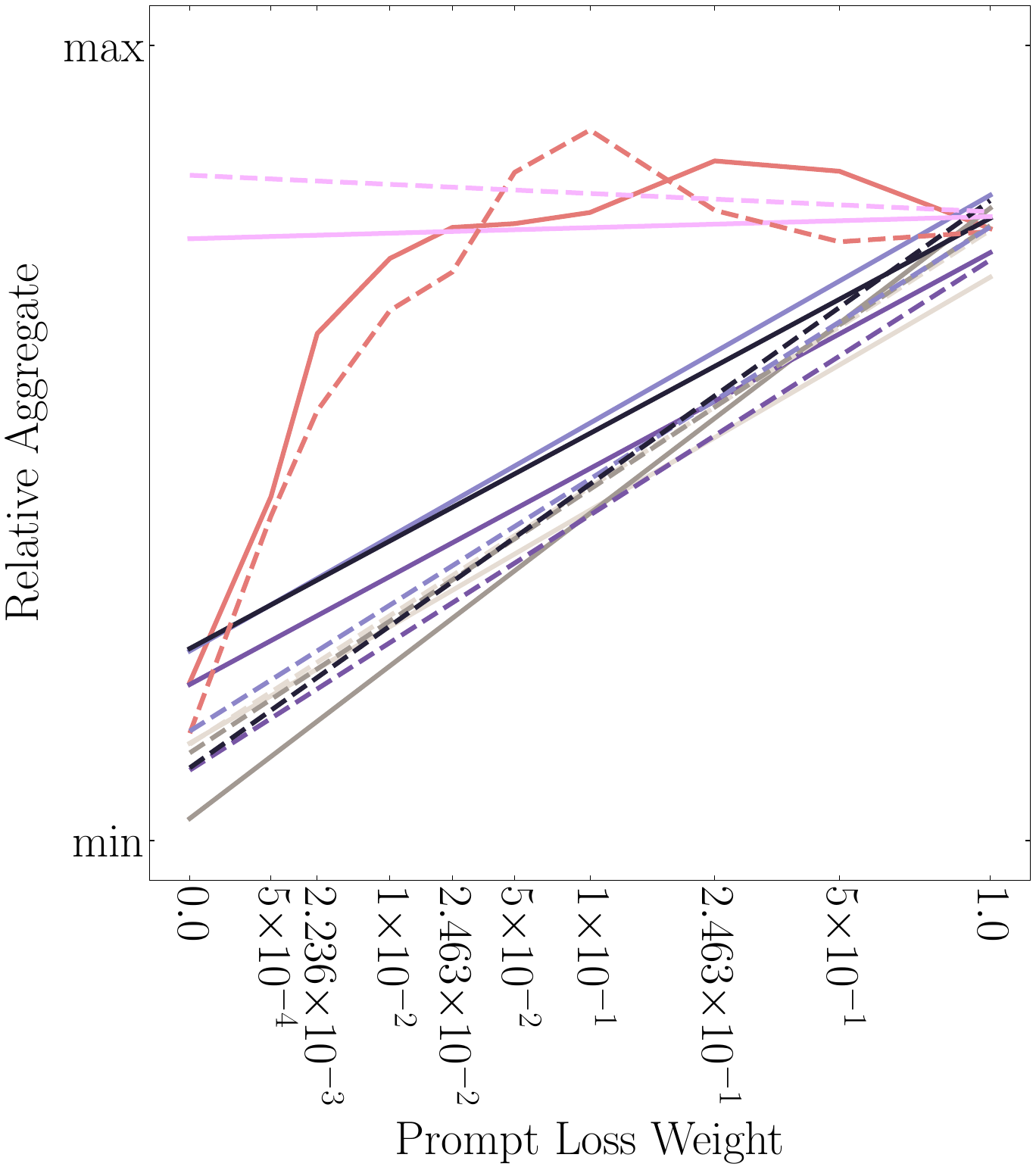}
        \vspace{-0.6cm}
        \caption{Relative Aggregate}
    \end{subfigure}
    \hspace{1cm}
    \begin{subfigure}[b]{.25\linewidth}
        \includegraphics[width=\linewidth]{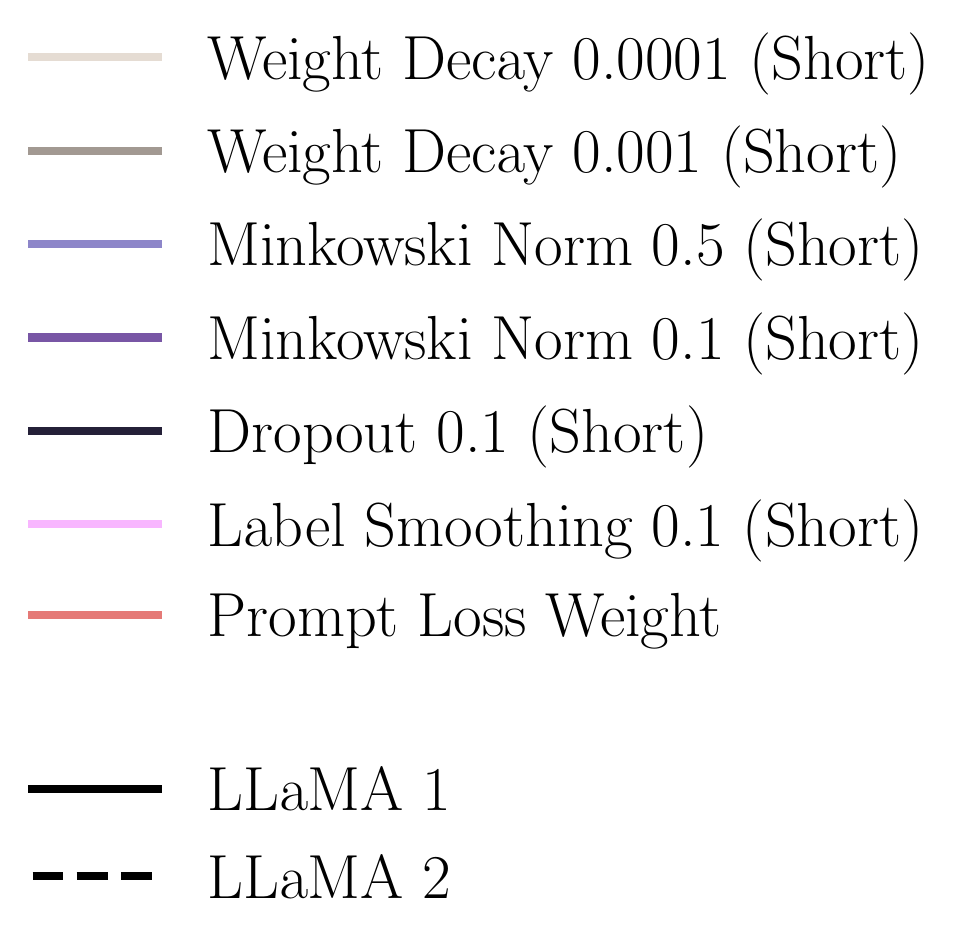}
        \vspace{-1.35cm}
    \end{subfigure}
    \vspace{0.6cm} \\
    \begin{subfigure}[b]{.3\linewidth}
        \includegraphics[width=\linewidth]{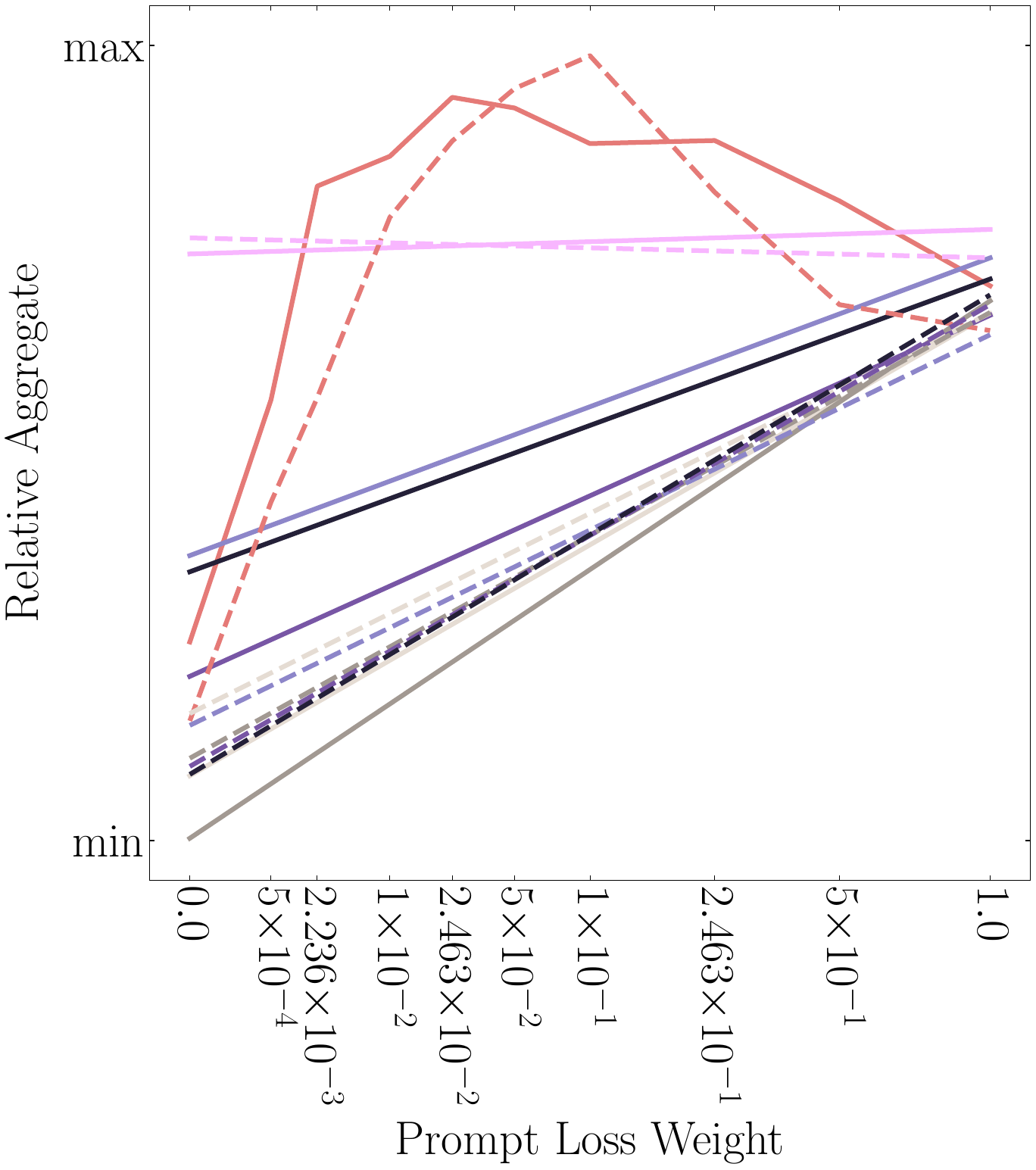}
        \vspace{-0.6cm}
        \caption{Relative Aggregate (Group I)}
    \end{subfigure}
    \hspace{0.01\linewidth}
    \begin{subfigure}[b]{.3\linewidth}
        \includegraphics[width=\linewidth]{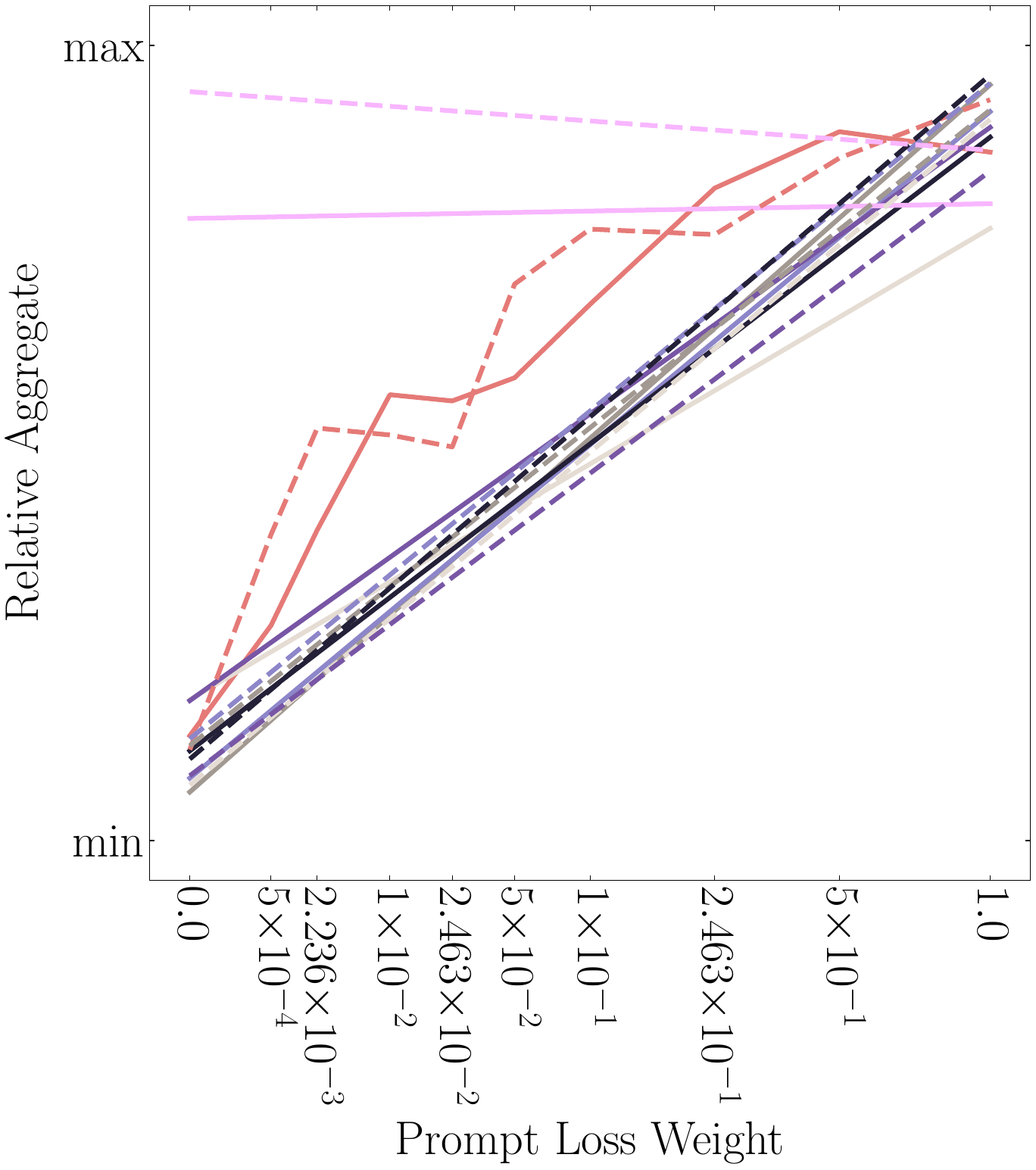}
        \vspace{-0.6cm}
        \caption{Relative Aggregate (Group II)}
    \end{subfigure}
    \caption{
    Comparison of PLW with other regularization techniques (calculated for PLW $=0$ and PLW $=1$).
    \textbf{(a)} The simple aggregate.
    \textbf{(b)} The combined relative aggregate shows that models fine-tuned with fractional PLW on AlpacaDataShort outperformed models fine-tuned with alternative regularizations.
    \textbf{(c)} Fractional PLW performance is most extreme for multiple choice and short-generation benchmarks (group I).
    \textbf{(d)} Performance of fractional PLW models on group II benchmarks is less pronounced, with PLW-optimized models performing slightly worse than several other alternative metrics.
    }
    \label{fig:short-reg}
\end{figure*}

\subsection{Regularization Comparison}\label{sec:app-reg}

For the first supplemental experiment, we wanted to investigate if PLW is necessary for fine-tuning on short-completion data or if another regularization technique could yield the same benefits.
As explained in section~\ref{sec:supp-reg}, we chose to examine four types of regularization in addition to PLW, intentionally not evaluating KL divergence-based regularization due to the difficulty of applying it to LLM SIFT.

Several aggregate scores for regularizations with example parameters are presented in figure~\ref{fig:short-reg}, and best scores for each type of regularization is presented in table~\ref{tab:short-reg} in the main paper.
Visualization of relative aggregate scores revealed that models fine-tuned with fractional PLW generated high scores on multiple-choice and short-generation benchmarks while long-generation benchmarks (AlpacaEval 1 and PandaLM) actually benefitted the most from alternative regularization methods.
However, the effect on the multiple choice and short generation benchmarks was relatively strong, and the combined relative aggregate also showed maximal values for models fine-tuned with fractional PLW.

Interestingly, most regularization methods performed better when coupled with PLW$=1$ than with PLW$=0$ except for label smoothing which performed marginally worse at PLW$=1$ than at PLW$=0$.

\subsection{Dataset Comparison}\label{sec:app-aug}

\begin{figure*}[t!]
    \begin{subfigure}[b]{.3\linewidth}
        \includegraphics[width=\linewidth]{images/augmentlong_nolegend_relativeagg.pdf}
        \vspace{-0.6cm}
        \caption{Original Datasets}
    \end{subfigure}
    \hspace{0.01\linewidth}
    \begin{subfigure}[b]{.3\linewidth}
        \includegraphics[width=\linewidth]{images/augmentshort_nolegend_relativeagg.pdf}
        \vspace{-0.6cm}
        \caption{Short Dataset Variants}
    \end{subfigure}
    \begin{subfigure}[b]{.20\linewidth}
        \includegraphics[width=\linewidth]{images/augment_agg_legend.pdf}
        \vspace{3cm}
    \end{subfigure}
    \\\\
    \vspace{0.3cm}
    \begin{subfigure}[b]{.3\linewidth}
        \includegraphics[width=\linewidth]{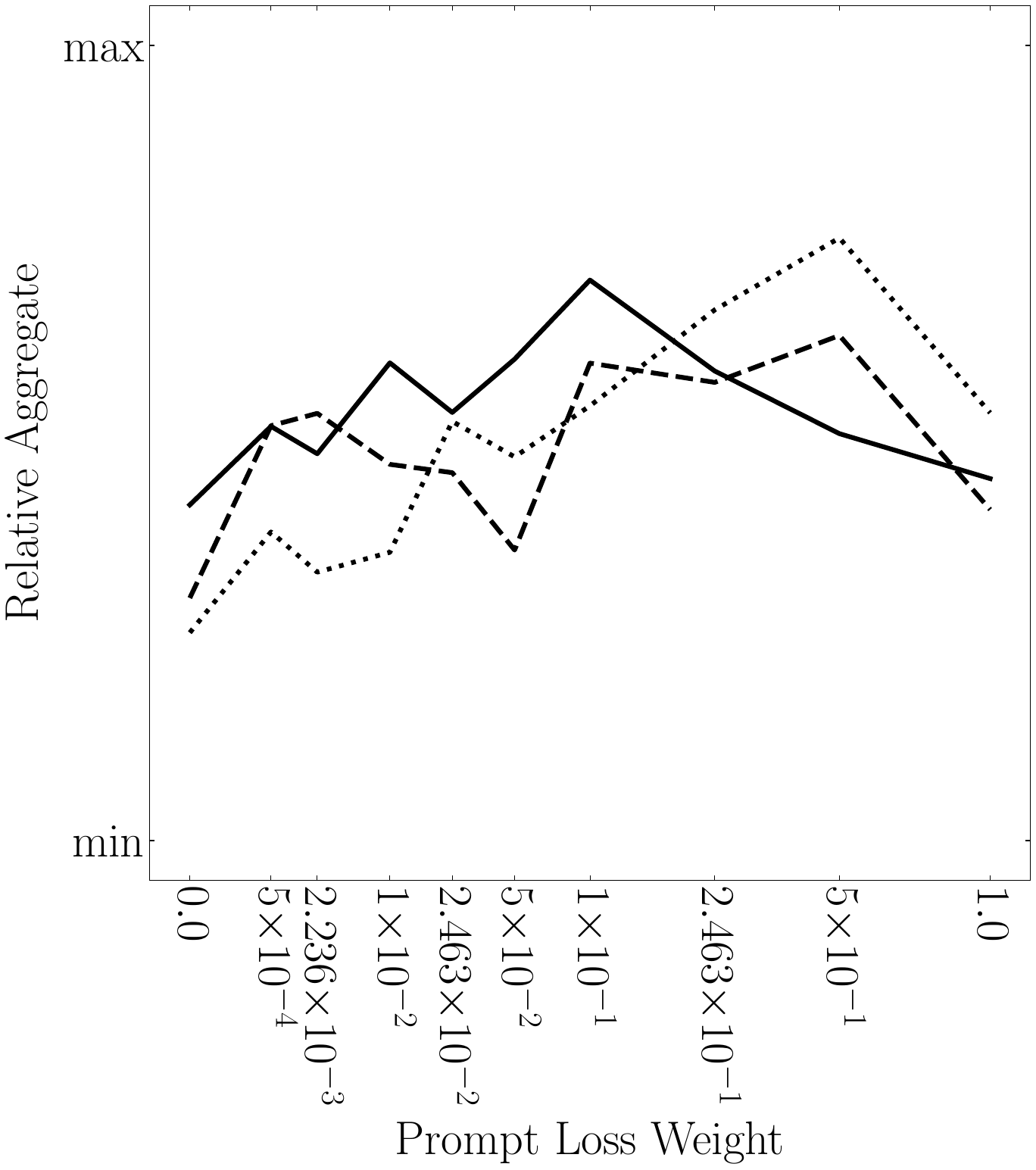}
        \vspace{-0.6cm}
        \caption{UltraFeedbackBinarizedClean}
    \end{subfigure}
    \hspace{0.01\linewidth}
    \begin{subfigure}[b]{.3\linewidth}
        \includegraphics[width=\linewidth]{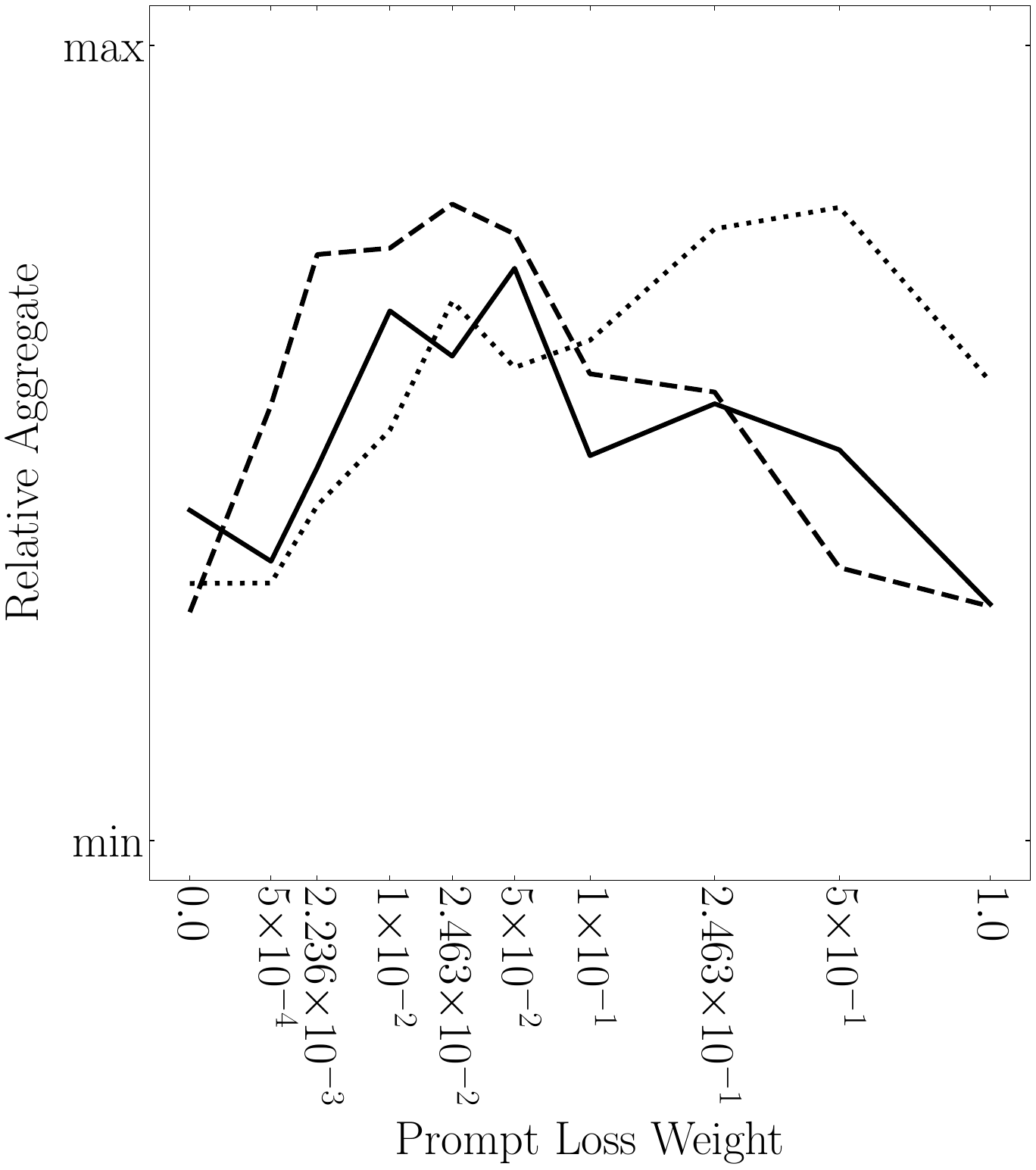}
        \vspace{-0.6cm}
        \caption{UltraFeedbackBinarizedShort}
    \end{subfigure}
    \\\\
    \begin{subfigure}[b]{.3\linewidth}
        \includegraphics[width=\linewidth]{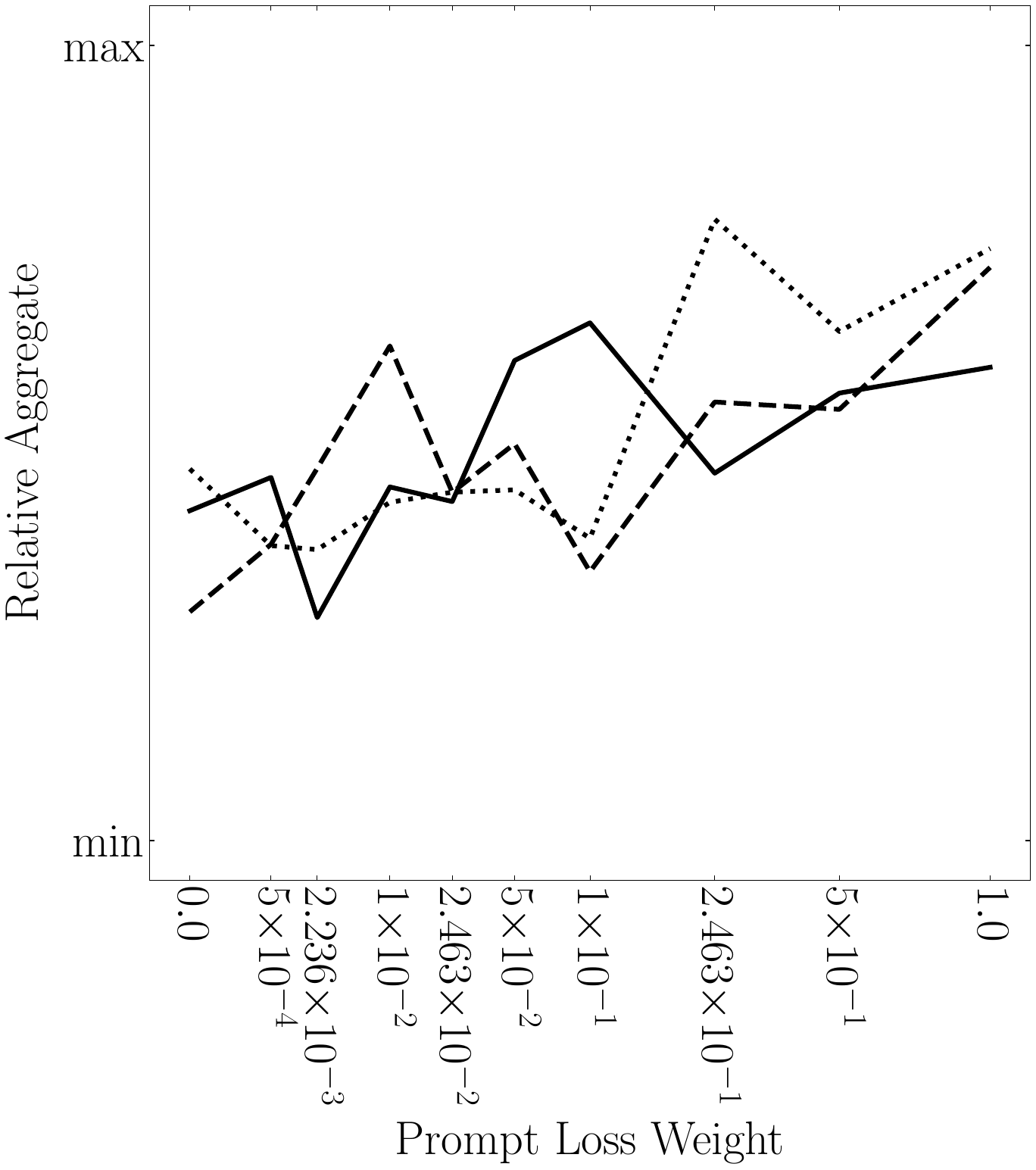}
        \vspace{-0.6cm}
        \caption{DatabricksDolly}
    \end{subfigure}
    \hspace{0.01\linewidth}
    \begin{subfigure}[b]{.3\linewidth}
        \includegraphics[width=\linewidth]{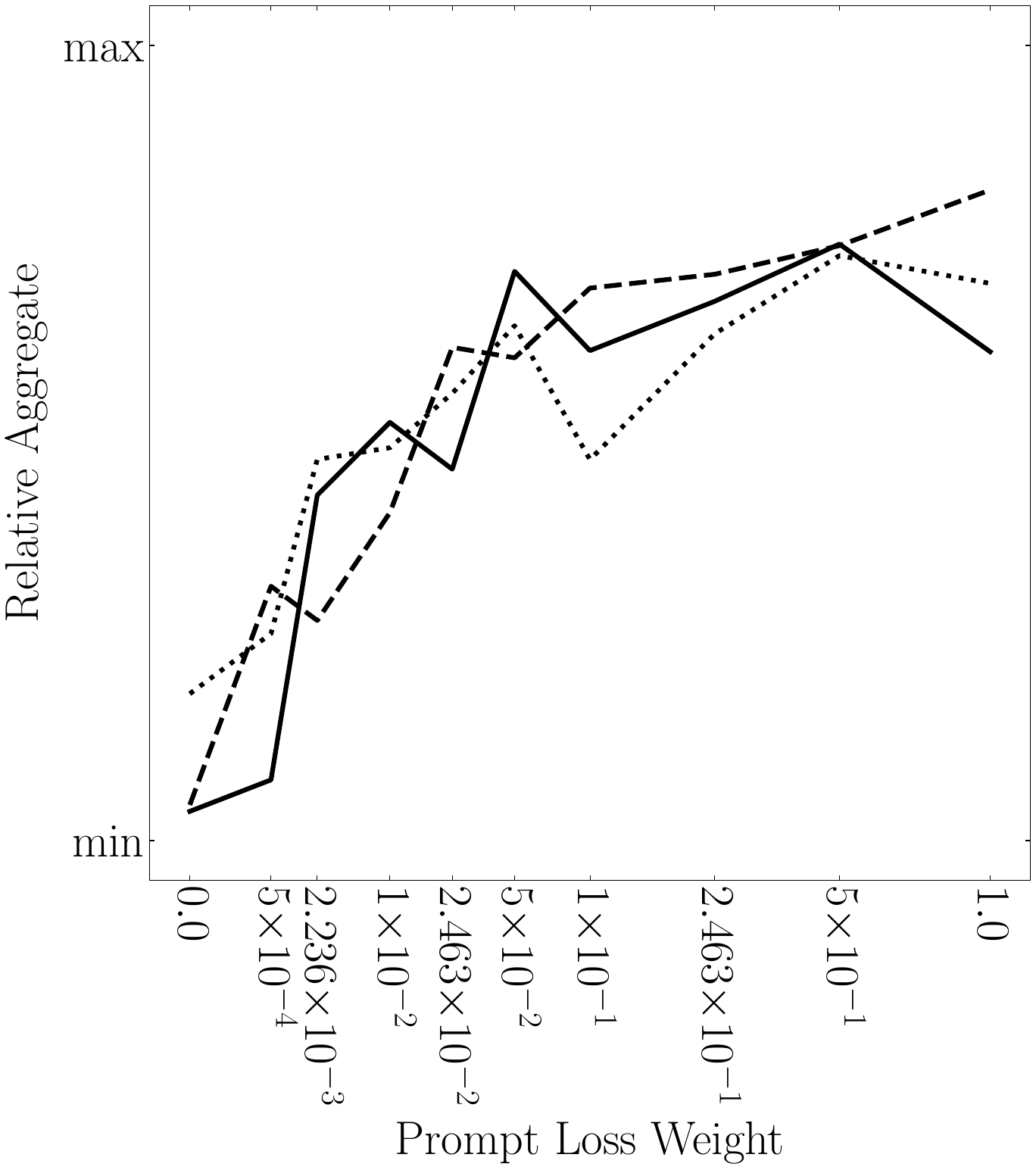}
        \vspace{-0.6cm}
        \caption{DatabricksDollyShort}
    \end{subfigure}
    \caption{
    Relative aggregate scores for models fine-tuned on alternative instruction datasets.
    }
    \label{fig:app-aug}
\end{figure*}

In the second supplemental experiment, we wanted to explore if the relationship between PLW and model performance on downstream tasks measured for AlpacaDataShort models existed for other fine-tuning datasets as well.
See additional visualizations in figure~\ref{fig:app-aug}. 

\vfill

\clearpage

\section{Optimal Prompt Loss Weight}
\label{sec:app-gam}

\begin{figure*}[bt!]
    \centering
    \begin{subfigure}[b]{.32\linewidth}
        \includegraphics[width=\linewidth]{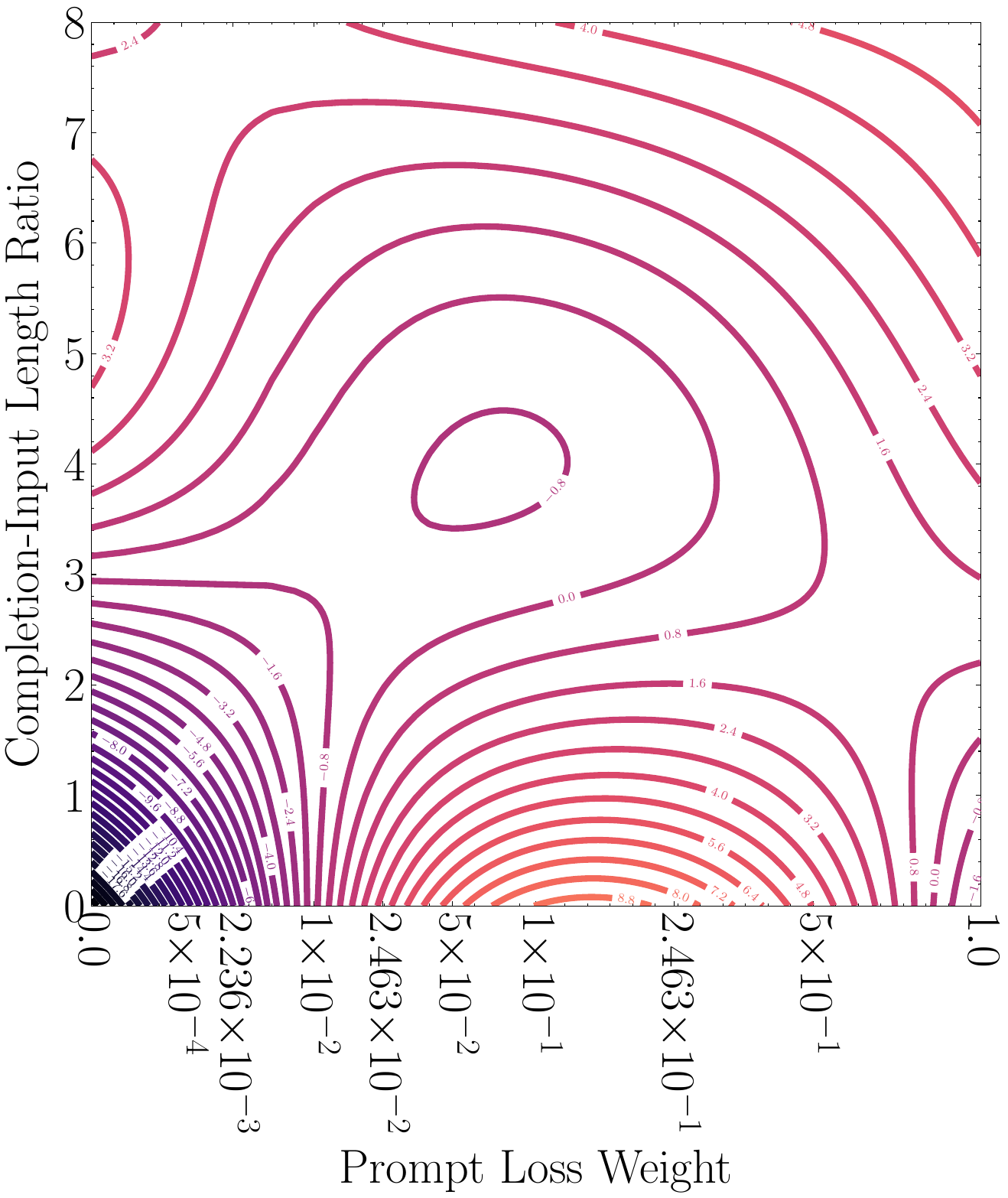}
        \caption{``All'' Prediction}
        \label{fig:gam-all}
    \end{subfigure}
    \begin{subfigure}[b]{.32\linewidth}
        \includegraphics[width=\linewidth]{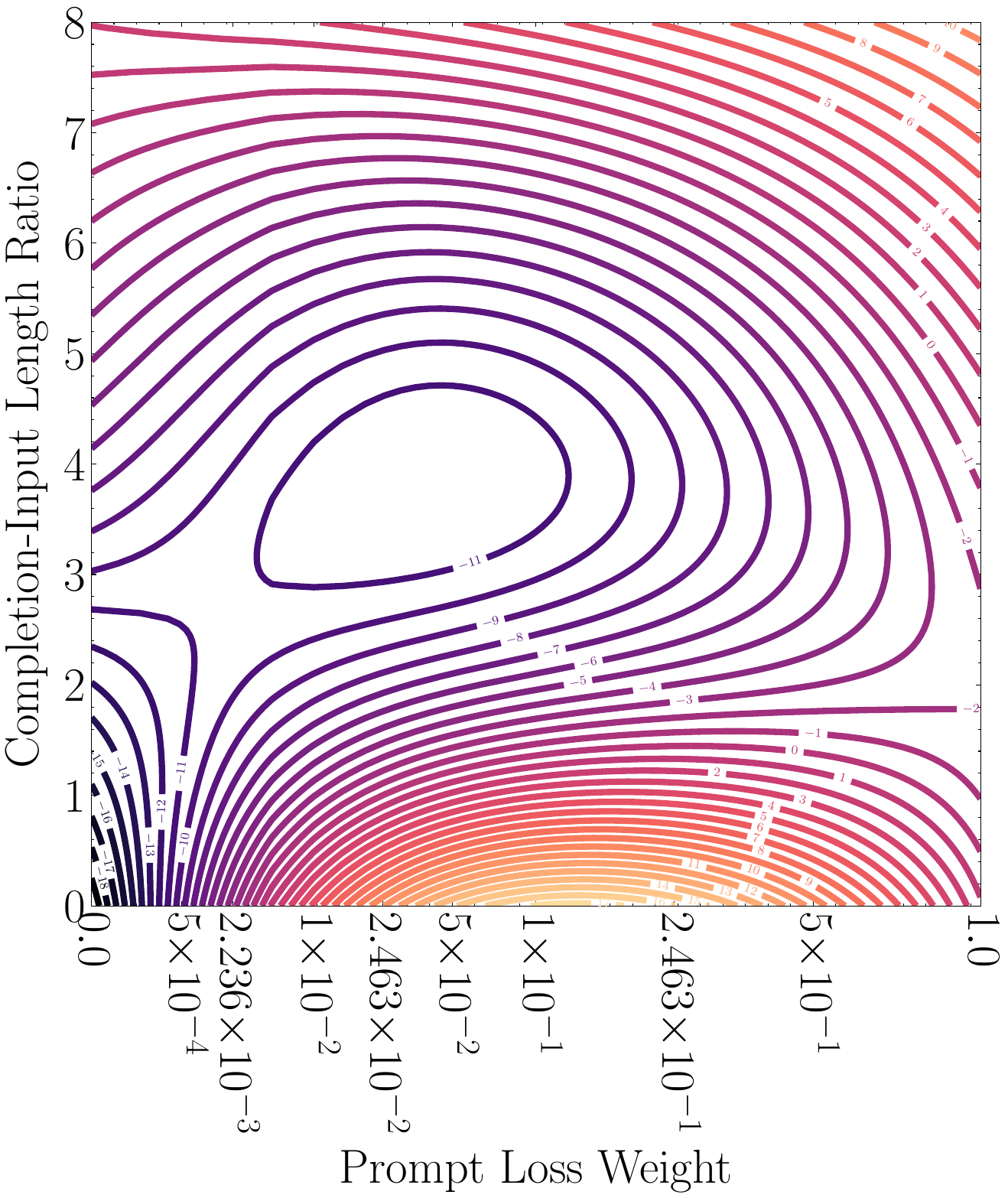}
        \caption{``MC'' Prediction}
        \label{fig:gam-mc}
    \end{subfigure}
    \begin{subfigure}[b]{.32\linewidth}
        \includegraphics[width=\linewidth]{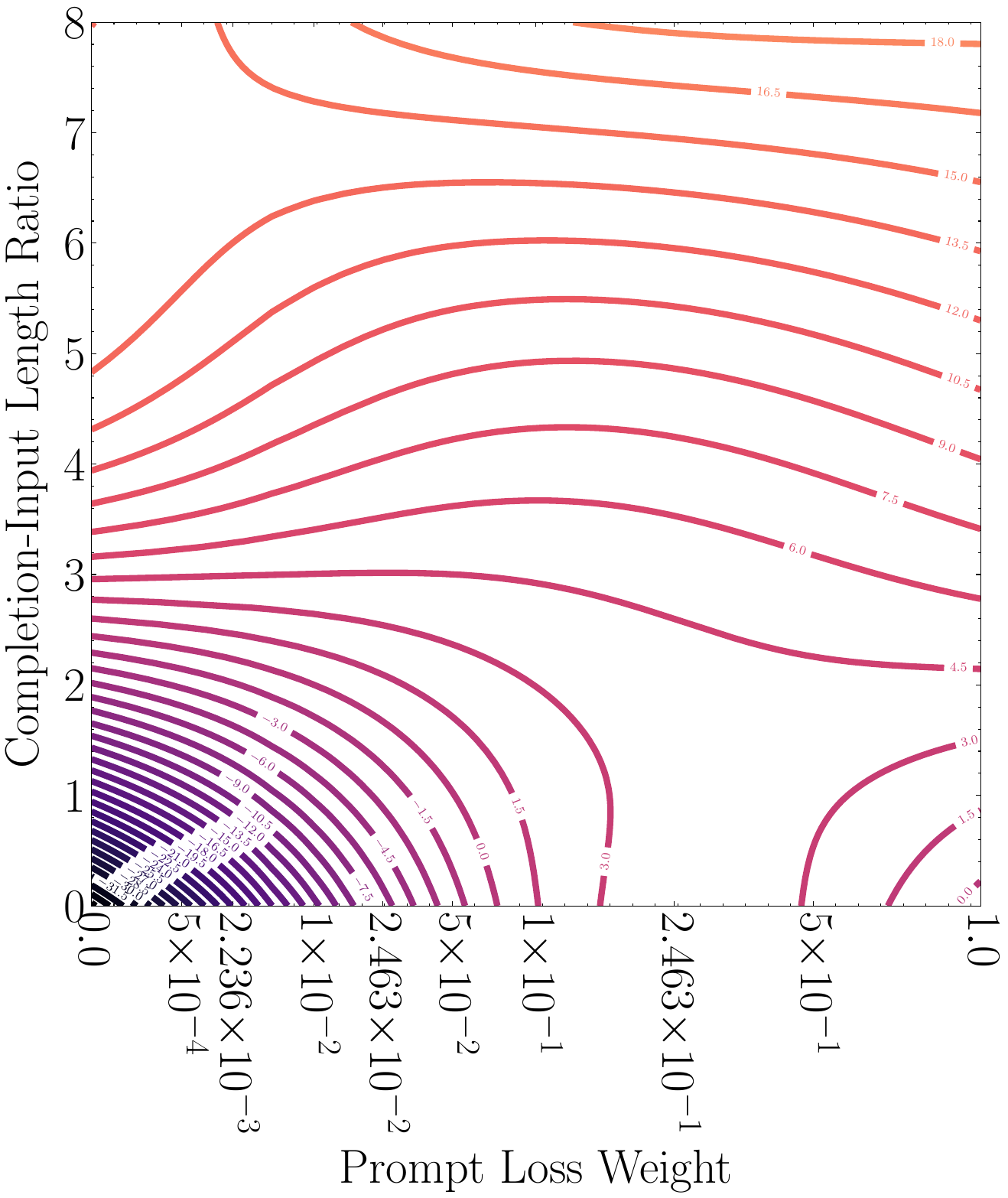}
        \caption{``Gen'' Prediction}
        \label{fig:gam-long}
    \end{subfigure}
    \caption{
    Best viewed digitally for improved resolution.
    }
    \label{fig:gam}
\end{figure*}

In the main regression experiment, we showed that PLW is an important hyperparameter when fine-tuning on short-completion data but is effectively irrelevant when using long-completion data.
In this appendix, we present several models for predicting an optimal PLW given a dataset's $R_g$.
These models are based on the AlpacaData dataset, AlpacaDataCleaned dataset, and several modified versions of AlpacaDataCleaned and therefore should be seen as an exercise rather than an authoritative reference on optimal PLW weights.

We first repeated our SIFT experiments on two additional datasets: AlpacaDataMedium and AlpacaDataTiny to increase coverage of the parameter space.
AlpacaDataMedium and AlpacaDataTiny have $R_g$ values of 1.0 and 0.042, respectively, and were generated using prompt inversion (see Appendix~\ref{sec:app-short}) but selecting instances to modify in order to approach target $R_g$ values.

We then fit several generalized additive model (GAMs) with a tensor smooth for the interaction between PLW and $R_g$.
GAMs offer more flexible modeling than linear models but at the expense of interpretability.
We fit our models using the \textbf{\textsf{R}} library \texttt{mgcv} and the following equation:\\
``$\texttt{score \textasciitilde{} te(w, r, k=3) + factor(b)}$'',
where \texttt{te} is a full tensor product smooth, \texttt{w} is the untransformed PLW parameter, \texttt{r} is the $R_g$, \texttt{k} is the number of splines, and \texttt{b} is the benchmark task.

\begin{table}[b!]
    \centering\small
    \begin{tabular}{cccc}
        \toprule
        \multirow{2}{*}{$R_g$} & \multicolumn{3}{c}{Optimal PLW} \\
        \cmidrule(lr){2-4}
        & All & MC & Gen \\
        \midrule
        8.0 & 1.000* & 1.000* & 0.654 \\
        7.5 & 1.000* & 1.000* & 1.000* \\
        7.0 & 1.000* & 1.000* & 1.000* \\
        6.5 & 1.000* & 1.000* & 1.000* \\
        6.0 & 1.000* & 1.000* & 0.000* \\
        5.5 & 1.000* & 1.000* & 0.000* \\
        5.0 & 0.000* & 1.000* & 0.000* \\
        4.5 & 0.000* & 1.000* & 0.000* \\
        4.0 & 1.000* & 1.000* & 0.000* \\
        3.5 & 1.000* & 1.000* & 0.000* \\
        3.0 & 1.000* & 1.000* & 1.000* \\
        2.5 & 1.000* & 1.000* & 1.000* \\
        2.0 & 0.239 & 1.000* & 0.679 \\
        1.5 & 0.183 & 0.278 & 0.385 \\
        1.0 & 0.155 & 0.183 & 0.321 \\
        0.5 & 0.155 & 0.155 & 0.292 \\
        \bottomrule
    \end{tabular}
    \caption{
    Optimal prompt loss weight (PLW) per completion-prompt length ratio $R_g$ on all benchmarks (``All''); multiple choice benchmarks (``MC''); and the combination of TruthfulQA-Gen, Alpaca Eval 1, and PandaLM benchmarks (``Gen'').
    Predictions are based on the ratio-PLW interaction term of fitted generalized additive models.
    \\
    *The difference between the maximum and minimum predicted values at this ratio is less than 5\% of the score range.
    }
    \label{tab:gam-all}
\end{table}

Using the fitted \texttt{w}-\texttt{r} interaction, we then estimated optimal PLW value for a given completion-prompt ratio $R_g$.
See figure~\ref{fig:gam-all} for a visualization of a GAM fitted on all benchmark tasks.

Roughly, the fitted interaction term recommended using PLW $=0.155$ for small completion-prompt length ratios ($R_g\le 1$) and up to PLW $=0.278$ for a $R_g = 1.5$ for optimal performance across all tasks.
This prediction is close to our regression-predicted value of 0.242.
The interaction term also confirms our observations that PLW is less important for data with relatively long completions.

Since the relationship between PLW and benchmark performance depends heavily on the type of benchmark task, we also fit GAMs for an aggregate of multiple choice benchmark scores (labeled ``MC'') and generation benchmark scores (labeled ``Gen'').
We found that the translation benchmarks contributed little to the predictive power of the fitted GAMs and while their scores are included in the ``All'' GAM, we did not include them when fitting the ``Gen'' GAM.
See figures~\ref{fig:gam-mc} and \ref{fig:gam-long} for contour plots for the ``MC'' and ``Gen'' benchmarks, respectively.

Also see table~\ref{tab:gam-all} for a list of GAM-based optimal PLWs over a range of completion-prompt ratios.
Again, predicted optimal PLWs confirmed the conclusions from our regression analysis in section~\ref{sec:regression}.

\clearpage
\section{Reproducibility}
\label{sec:app-reproduce}

This section provides technical details on all experiments and benchmarks for transparency and to encourage reproduction of results.
To help with reproducibility, we will also uploaded the fine-tuned models, test generation outputs, and our modified datasets to the HuggingFace Hub and can be accessed at \url{https://huggingface.co/collections/mathewhe/plw-66fe40fac6e586d8435bd563}.
Note that unless specified otherwise, default parameters were used for all training and testing.

\subsection{Model Fine-Tuning}
Model fine-tuning was performed with the Stanford Alpaca GitHub repository at \url{https://github.com/tatsu-lab/stanford_alpaca/tree/761dc5b}.

To experiment with prompt loss weight, we modified HuggingFace's Transformers library to allow specifying a \texttt{loss\_weights} parameter for \texttt{LlamaForCausalLM}'s forward method.

We used the following commit of Transformers \url{https://github.com/huggingface/transformers/tree/3b7675b}.

All models were trained on a single four A100 80GB node and we used the first set of hyperparameters recommended in the Fine-tuning subsection of Stanford Alpaca's README.md file, except for the three experimental variables: pretrained model, prompt loss weight, and training dataset.

AlpacaData is available from the Stanford Alpaca repository.
AlpacaDataCleaned can be found at \url{https://github.com/gururise/AlpacaDataCleaned/tree/791174f} and is labeled ``alpaca\_data\_cleaned.json''.
As noted above, AlpacaDataShort can be accessed at \url{https://huggingface.co/collections/mathewhe/plw-66fe40fac6e586d8435bd563}.

\subsection{Model Evaluation}
\label{sec:app-reproduce-eval}

We used three evaluation frameworks: EleutherAI's Language Model Evaluation Harness (EEH), AlpacaEval 1, and PandaLM.

In an effort to match the current HuggingFace Open LLM leaderboard, we evaluated ARC Challenge, TruthfulQA-MC2, WinoGrande, and PIQA on the same EEH commit that the HuggingFace leaderboard uses: \url{https://github.com/EleutherAI/lm-evaluation-harness/tree/b281b09}
We also matched the number of shots with the number used for the HuggingFace leaderboard for ARC Challenge, TruthfulQA-MC2, and WinoGrande.

TruthfulQA-Gen and all translation tasks were evaluated using a more recent commit at \url{https://github.com/EleutherAI/lm-evaluation-harness/tree/b93c3bc}.
We modified the translation tasks at this commit to include an appropriate prompt to support zero-shot translation.
These changes can be seen at \url{https://github.com/mathewhuen/plw_lm-evaluation-harness/compare/b93c3bc..1957d1a}.

Though version 2 of AlpacaEval has recently been released, we used version 1 from the following commit \url{https://github.com/tatsu-lab/alpaca_eval/tree/495b606}.
To use Mixtral 8x7B as an auto-evaluator for AlpacaEval 1, we modified the Guanaco-33b evaluator's config and prompt minimally to match Mixtral's format.
Models were evaluated on the default test set which can be found at \url{https://huggingface.co/datasets/tatsu-lab/alpaca_eval/blob/main/alpaca_eval.json}.
We plan on submitting a pull request with these additions in the near future.

For PandaLM, we used the commit at \url{https://github.com/WeOpenML/PandaLM/tree/eb758c4} and evaluated on version 1 of the default test set (found at ``data/testset-inference-v1.json'' in the PandaLM repository).

\subsection{Regression}

All statistical analysis and regression modeling was performed with \textbf{\textsf{R}}, version 4.3.0.
We used the \texttt{glmmTMB} library, version 1.1.8, to perform generalized linear mixed modeling (GLMM) and validated results with the same library and with \texttt{DHARMa}, version 0.4.6.

\subsection{Causal Mechanism}

Most of the analysis performed to shed light on the causal mechanism should version and implementation agnostic.
However, BLEU score implementations vary widely, and we used sacre BLEU to evaluate memorization of the training set.
We used Corpus BLEU from the sacreBLEU library at \url{https://github.com/mjpost/sacrebleu}.
Instead of a commit hash, we share the metric signature: \\
\texttt{``nrefs:1|case:mixed|eff:no|tok:13a|}
\texttt{smooth:exp|version:2.4.0''}

\subsection{Supplemental Experiments}

The first supplemental experiment used the same experimental setup and data from the main experiment.
Of the tested regularization methods, we used the weight decay implementation in PyTorch's AdamW optimizer, attention dropout implemented by the LlamaAttention module from Transformers, and label smoothing supported by the Trainer class from Transformers.
We manually implemented regularization based on the Minkowski distance by calculating the mean of the $p=1$ Minkowski distance between each pair of weight tensors from the PTLM and the trained model.

The second experiment introduced two new datasets: UltraFeedbackBinarizedCleaned and DatabricksDolly.
UltraFeedbackBinarizedCleaned can be found at \url{https://huggingface.co/datasets/allenai/ultrafeedback_binarized_cleaned/tree/f304ce5}.
And DatabricksDolly can be found at \url{https://huggingface.co/datasets/databricks/databricks-dolly-15k/tree/bdd27f4}.
The modified datasets UltraFeedbackShort and DatabricksDollyShort can be accessed at \url{https://huggingface.co/collections/mathewhe/plw-66fe40fac6e586d8435bd563}.

\subsection{Predictive Model}

We fit several generalized additive models (GAMs) in Appendix~\ref{sec:app-gam} using the mgcv library, version 1.9-1 and the same version of \textbf{\textsf{R}} as above, version 4.3.0.

\clearpage
\section{Artifact Licensing}
\label{sec:license}

We respected all licenses for artifacts and resources used in this research.
Please see table~\ref{tab:license} for an overview of primary resources and licenses.

\begin{table*}[hbt!]
    \centering\small
    \begin{tabular}{ccc}
        \toprule
        Resource & License & Application \\
        \midrule
        Transformers & Apache 2.0 & Model Training \\
        Stanford Alpaca & Apache 2.0 & Model Training \\
        AlpacaDataCleaned & Apache 2.0 & Model Training \\
        Ultrafeedback Binarized Cleaned & MIT & Model Training \\
        databricks-dolly-15k & CC BY-SA 3.0 & Model Training \\
        LLaMA 1 & LLaMA License & Pre-trained model weights \\
        LLaMA 2 & LLaMA 2 Community License & Pre-trained model weights \\
        LLaMA 3 & LLaMA 3 Community License & Pre-trained model weights \\
        Mixtral 8x7B & Apache 2.0 & Model Evaluation \\
        EleutherAI's LM Evaluation Harness & MIT & Model Evaluation \\
        AlpacaEval 1 & Apache 2.0 & Model Evaluation \\
        AlpacaEval Dataset & CC BY-NC 4.0 & Model Evaluation \\
        PandaLM & Apache 2.0 & Model Evaluation \\
        ARC Challenge & CC BY-SA 4.0 & Model Evaluation \\
        PIQA & AFL 3.0 & Model Evaluation \\
        TruthfulQA & Apache 2.0 & Model Evaluation \\
        WinoGrande & Apache 2.0 & Model Evaluation \\
        WMT 14 & No License & Model Evaluation \\
        WMT 16 & No License & Model Evaluation \\
        \bottomrule
    \end{tabular}
    \caption{
    Licenses for resources used in this research.
    }
    \label{tab:license}
\end{table*}

\end{document}